\documentclass[journal]{IEEEtran}
\usepackage{amsmath,amsfonts}
\usepackage{algorithmic}
\usepackage{algorithm}
\usepackage{array}
\usepackage[caption=false,font=normalsize,labelfont=sf,textfont=rm]{subfig}
\usepackage{textcomp}
\usepackage{stfloats}
\usepackage{url}
\usepackage{verbatim}
\usepackage{graphicx}
\usepackage{cite}
\usepackage{booktabs}
\usepackage{multirow}
\usepackage{makecell}
\usepackage{colortbl}
\usepackage{hyperref}
\hypersetup{hypertex=true,
colorlinks=true,
linkcolor=blue,
anchorcolor=blue,
citecolor=blue}

\usepackage[inkscapelatex=false]{svg}
\newcommand{\edit}[1]{\textcolor{black}{#1}}
\definecolor{Gray}{gray}{0.94}
\begin{document}

\title{MoPD: Mixture-of-Prompts Distillation for Vision-Language Models}

\author{Yang Chen, Shuai Fu, Yu Zhang, \IEEEmembership{Member,~IEEE}
\thanks{Yang Chen and Yu Zhang are with the Department of Computer Science and Engineering, Southern University of Science and Technology, Shenzhen 518055, China (e-mail: cheny2023@mail.sustech.edu.cn; yu.zhang.ust@gmail.com).}
\thanks{Shuai fu is with the School of Mathematical and Computer Sciences, University of Adelaide, Australia. (e-mail: fus.jayce@gmail.com).}
\thanks{Corresponding author: Yu Zhang.}

\thanks{This paper has supplementary downloadable material available at http://ieeexplore.ieee.org., provided by the authors.}
}



\maketitle

\begin{abstract}
Soft prompt learning methods are effective for adapting vision-language models (VLMs) to downstream tasks. Nevertheless, empirical evidence reveals that existing methods tend to overfit seen classes and exhibit degraded performance on unseen classes. This limitation is due to the inherent bias in the training data towards the seen classes. To address this issue, we propose a novel soft prompt learning method, named Mixture-of-Prompts Distillation (MoPD), which can effectively transfer useful knowledge from hard prompts manually hand-crafted (a.k.a. teacher prompts) to the learnable soft prompt (a.k.a. student prompt), thereby enhancing the generalization ability of soft prompts on unseen classes. Moreover, the proposed MoPD method utilizes a gating network that learns to select hard prompts used for prompt distillation. Extensive experiments demonstrate that the proposed MoPD method outperforms state-of-the-art baselines, especially on unseen classes.
\end{abstract}


\section{Introduction}
\IEEEPARstart{W}{ith} large-scale training data of image-text pairs, large pre-trained vision-language models (VLMs) such as CLIP \cite{clip}, BLIP \cite{blip}, Flamingo \cite{flamingo}, and LLaVA \cite{llava} possess a remarkable zero-shot generalization ability and address various downstream vision tasks, such as image classification \cite{clip,ic_tmm1,ic_tmm2}, object detection \cite{zhong2022}, segmentation \cite{wang2022}, visual question answering \cite{blip, flamingo,vqa_tmm3,vqa_tmm4}, and visual reasoning \cite{llava}. 
Some soft prompt learning methods, such as CoOp \cite{coop} and CoCoOp \cite{cocoop}, are built on VLMs. Those methods incorporate learnable prompting vectors into the text encoder or decoder while keeping the pre-trained model fixed, enabling few-shot fine-tuning of VLMs. These learnable prompts act as data-efficient learners and enable large pre-trained VLMs to perform better on a variety of downstream tasks.
\begin{figure}[htbp]
    \centering
    \includegraphics[width=0.9\linewidth]{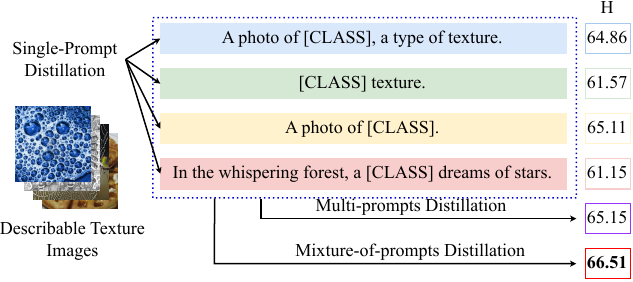}
    \caption{Comparison among single-prompt distillation, multi-prompts distillation, and the proposed MoPD method, where single-prompt distillation is to distill knowledge from a single hard prompt to the soft prompt and multi-prompts distillation is to distill knowledge from multiple hard prompts to the soft prompt. 
    `H' denotes the harmonic mean accuracy.
 }
    \label{fig_intro}
\end{figure}

Recently, some works (e.g., ProGrad \cite{prograd} and KgCoOp \cite{kgcoop}) have shown that soft prompt learning methods tend to overfit the base (seen) classes and generalize poorly on the new (unseen) classes, while VLMs with hard prompts (e.g., a photo of a [CLASS], where [CLASS] denotes the placeholder for the class name) can achieve good zero-shot recognition performance on new classes. 
To utilize the useful knowledge captured in the pre-trained VLMs through hard prompts, ProGrad regulates the gradient direction to mitigate potential conflicts between hard prompts and soft prompts. KgCoOp mitigates the risk of knowledge loss by minimizing the embedding discrepancy between soft prompts and hard prompts. \edit{Similarly, to align with pre-trained CLIP's generalization, LASP \cite{lasp} maximizes the probability of the learned prompts being correctly classified with respect to pre-defined hard prompts.}

However, the aforementioned methods solely rely on a single hard prompt, without considering the use of multiple hard prompts to enhance the learning of soft prompts. In fact, compared with a single hard prompt, a pool with multiple hard prompts may provide more useful knowledge for prompt-based models, and such a pool is commonly available. Besides, knowledge distillation techniques \cite{hinton2015, romero2015, furlanello2018, phuong2019} are an effective approach to transfer knowledge, but they are unexplored in existing soft prompt learning methods for VLMs. 

Based on the above considerations, in this paper we perform the first investigation on the effectiveness of multi-prompts distillation for VLMs by comparing with single-prompt distillation, where the distillation is from a hard prompt (a.k.a. teacher prompt) to the soft prompt (a.k.a. student prompt). According to the results shown in Fig. \ref{fig_intro}, we can see that simultaneously distilling the knowledge of multiple hard prompts can achieve better performance when compared with the single hard prompt, which verifies the effectiveness of the use of multiple hard prompts. 
However, the multi-prompts distillation may be vulnerable to the unrelated and noisy content of hard prompts. 
Moreover, it is challenging yet beneficial to find an instance-specific teacher prompt for each image due to the varying prompt preferences across different images \cite{cocoop}.

To solve these issues, we propose a \textbf{M}ixture-\textbf{o}f-\textbf{P}rompts \textbf{D}istillation (\textbf{MoPD}) method.
Specifically, the proposed MoPD method utilizes a trainable gating network to select one or more instance-specific hard prompts from a pool of hard prompts, guiding the learning of the soft prompt, where the gating network assigns weights to teacher prompts based on image features during the prompt learning stage. 
Extensive experiments on 11 datasets demonstrate that the proposed MoPD method outperforms state-of-the-art baseline methods under several settings (i.e., base-to-new, few-shot classification, domain generalization, \edit{and cross-dataset evaluation}).

The main contributions of this work are three-fold.
\begin{itemize}
\item We are the first to introduce instance-specific prompt distillation for VLMs to transfer useful knowledge from hard prompts to soft prompts and validate that distilling multiple hard prompts is more effective compared to a single hard prompt.
\item We propose Mixture-of-Prompts Distillation for VLMs, which introduces a gating network to select suitable hard prompts for effectively transferring useful knowledge to the soft prompt.
\item Extensive experimental results demonstrate the superior performance of MoPD. 
Moreover, our experiments have substantiated the robustness of MoPD to noisy prompts within the hard prompt pool.
\end{itemize}

\section{Related Work}
\subsection{Vision-Language Models}
In recent years, large pre-trained VLMs~\cite{clip, blip, align, flamingo, llava} have shown great impacts in computer vision. A representative work is CLIP \cite{clip}, which jointly trains an image encoder built on a Vision Transformer (ViT) \cite{vit} or ResNet \cite{resnet} and a transformer-based text encoder with 400 million image-text pairs via a contrastive objective, showing excellent zero-shot generalization capability. ALIGN \cite{align} is closely related to CLIP, with the difference that ALIGN uses uncleaned training data, so the scale of the collected training data can be larger but noisier. In this paper, we use CLIP as the foundation model.

\subsection{Prompting}
Prompting techniques \cite{prompting1, prompting2} enable the effective utilization of knowledge in pre-trained models by reformulating the downstream task into a ``fill-in-the-blank'' cloze test format through prompts. In VLMs, prompts are extensively employed to efficiently adapt the model to various downstream tasks while keeping a small number of training parameters. For instance, CLIP converts a conventional prediction task into the task of associating captions with images, and enables zero-shot transfer to downstream tasks by designing hard prompts like ``a photo of a [CLASS]''. However, designing prompts for different downstream tasks is time-consuming and always requires domain expertise, and the designed prompts are not guaranteed to be optimal.

\subsection{Soft Prompt Learning}
To solve the above issue in prompting, soft prompt learning methods are proposed to learn a group of prompting vectors while fixing parameters in the backbone. For example, CoOp \cite{coop} utilizes learnable prompting vectors to replace the prompting words like ``a photo of a'' and adapts to downstream tasks. 
\edit{SgVA-CLIP \cite{sgva} combines prompt learning and adapter \cite{adapter}, producing more discriminative visual information by comprehensively considering both adapted visual feature space and pre-trained cross-modal feature space, thus enhancing few-shot image classification.} However, \edit{the previous methods} tend to overfit seen classes, thereby limiting their ability to generalize to unseen classes within the same dataset. To address the issue of weak generalization ability, CoCoOp \cite{cocoop} trains a lightweight neural network to generate input-conditional vectors for each image, collaborating with prompting vectors to dynamically generate prompting vectors. As discussed before, ProGrad, KgCoOp, and \edit{LASP} aim to use a hard prompt to guide the learning of the soft prompt in different ways. \edit{However, using only textual prompt learning as the aforementioned methods did is sub-optimal. To alleviate that, MaPLe \cite{maple} introduces a branch-aware multi-modal prompt learning method that simultaneously tunes vision and language branches together by sharing soft prompts across both modalities. Unlike MaPLe, PromptSRC \cite{promptsrc} trains independent soft prompts for text and image encoders, and introduces a self-regularization framework to learn general features from the frozen CLIP. Similarly, CoPrompt \cite{coprompt} employs consistency constraints for prompt learning to enhance the generalization and adopts a large language model to generate more descriptive sentences. MetaPrompt \cite{metaprompt} enhances the generalization by utilizing gradients from meta-test subtasks to regularize soft prompts.}

To the best of our knowledge, prompt distillation for VLMs remains unexplored. Given the potential of this technique to transfer knowledge, this paper investigates the efficacy of single-prompt distillation and multi-prompt distillation in the context of VLMs. 

\section{Methodology}
In this section, we first briefly review CLIP and CoOp, and then introduce the proposed MoPD method in detail.

\begin{figure*}[t]
  \centering
  \includegraphics[width=0.8\linewidth]{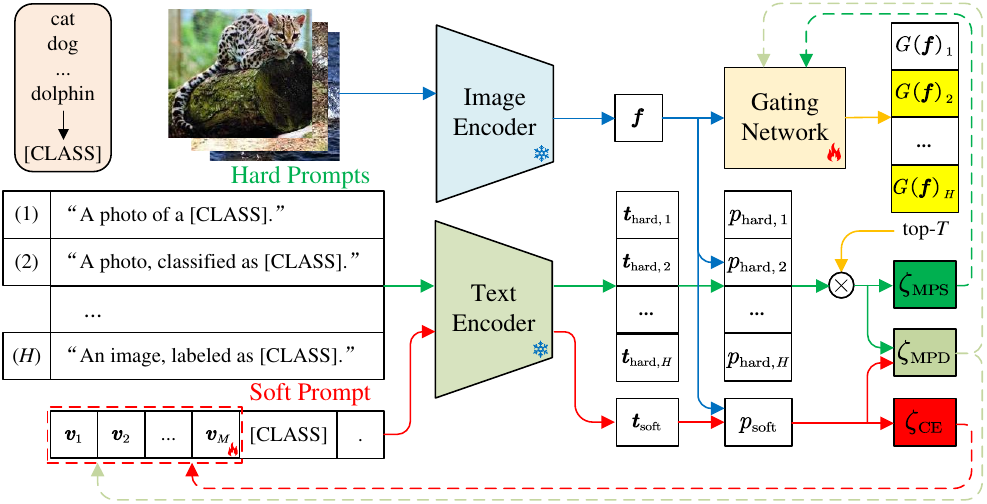}
  \caption{Overview of the proposed MoPD method, where solid arrows represent forward propagation processes and dashed arrows indicate backward propagation processes.}
  \label{framework}
\end{figure*}
\subsection{Preliminaries}
CLIP is pre-trained on about 400 million image-text pairs, resulting in its remarkable zero-shot image recognition capability. CLIP comprises two encoders: an image encoder, which is responsible for extracting visual information, and a text encoder, which is used to extract textual information. 

Given a set of class names with a total number of $C$ classes, a hard prompt like ``a photo of a [$\mathrm{CLASS}$]'' is \edit{first tokenized into discrete tokens and converted into word embeddings. These embeddings are then fed into the text encoder to generate textual embedding $\boldsymbol{t}_{\mathrm{hard}}\in \mathbb{R}^{C\times d}$, where $d$ is the dimension (i.e., 512 for CLIP).}
The visual embedding $\boldsymbol{f}$ is extracted by the image encoder for an image $\boldsymbol{x}$. The prediction probability of image $\boldsymbol{x}$ belonging to the $y$-th class is formulated as
\begin{equation}
p_{\mathrm{hard}}\left( y|\boldsymbol{x} \right) =\frac{\exp \left( \cos \left( \boldsymbol{t}_{\mathrm{hard}}^{y},\boldsymbol{f} \right) /\tau \right)}{\sum\nolimits_{c=1}^C{\exp \left( \cos \left( \boldsymbol{t}_{\mathrm{hard}}^{c},\boldsymbol{f} \right) /\tau \right)}},
  \label{eq_p_hard}
\end{equation}
where $\tau$ is a temperature parameter, $\cos \left( \cdot ,\cdot \right)$ denotes the cosine similarity, \edit{and $\boldsymbol{t}_{\mathrm{hard}}^{y}$ denotes the textual embedding of the $y$-th class generated by the hard prompt.} 

However, hard prompts (e.g., ``a photo of a [$\mathrm{CLASS}$]''), which are straightforward for humans, may not be optimal for VLMs. To enhance the adaptability of prompts for various downstream tasks, the CoOp method replaces such prompting words with learnable prompting vectors, which are called soft prompts. This soft prompting approach has demonstrated improved performance on multiple downstream datasets compared with using hard prompts \cite{clip,align}.
Specifically, the soft prompt consists of learnable prompting vectors of length $M$, where each prompting vector has the same dimension as the word embedding in VLMs, and a class name. Hence, the soft prompt can be defined as $\left\{ \boldsymbol{v}_1, \boldsymbol{v}_2, \ldots, \boldsymbol{v}_M, \left[ \mathrm{CLASS} \right] \right\} $. The textual embedding can be obtained by feeding the soft prompt into the text encoder, and it is denoted by \edit{$\boldsymbol{t}_{\mathrm{soft}}\in \mathbb{R}^{C\times d}$}. Then the prediction probability of image $\boldsymbol{x}$ belonging to the $y$-th class is formulated as
\begin{equation}
p_{\mathrm{soft}}\left( y|\boldsymbol{x} \right) =\frac{\exp \left( \cos \left( \boldsymbol{t}_{\mathrm{soft}}^{y},\boldsymbol{f} \right) /\tau \right)}{\sum\nolimits_{c=1}^C{\exp \left( \cos \left( \boldsymbol{t}_{\mathrm{soft}}^{c},\boldsymbol{f} \right) /\tau \right)}}\edit{,}
  \label{eq_p_soft}
\end{equation}
\edit{where $\boldsymbol{t}_{\mathrm{soft}}^{y}$ denotes the textual embedding of the $y$-th class generated by the learnable soft prompt.}
The soft prompt can be learned by minimizing the cross-entropy loss as
\begin{equation}
\zeta _{\mathrm{CE}}=-\sum_{(\boldsymbol{x},y)\in \mathcal{D}}{{\ln p_{\mathrm{soft}}\left( y|\boldsymbol{x} \right)}},
  \label{eq_sce}
\end{equation}
where $\mathcal{D}$ denotes the training set. Note that the image encoder and text encoder are frozen during the process of soft prompt learning.

\subsection{Mixture-of-Prompts Distillation}
\label{sec_MoPD}
Besides a frozen image encoder and a frozen text encoder for extracting visual and textual information as well as a pool of hard prompts, MoPD consists of a soft prompt and a trainable gating network to select instance-specific hard prompts for each image to guide the process of soft prompt learning. The overall framework of MoPD is shown in Fig. \ref{framework}.

\subsubsection{Single-Prompt Distillation (SiPD)} 
Here we first introduce how to perform distillation to guide the learning of a soft prompt based on a hard prompt, which will set the stage for the following introduction of MoPD. 

Knowledge distillation aims at transferring the knowledge from a teacher model to a student model. Similarly, we introduce prompt distillation as a technique to enhance the generalization ability of soft prompts based on hard prompts. In prompt distillation, a hard prompt serves as the teacher prompt, guiding the learning of the soft prompt, which acts as the student prompt. 
Considering Eqs. \eqref{eq_p_hard} and \eqref{eq_p_soft}, the prediction probability distributions of the hard and soft prompts can be formulated as $p_{\mathrm{hard}}( \cdot|\boldsymbol{x} )$ and $p_{\mathrm{soft}}( \cdot|\boldsymbol{x} )$, respectively. Then the prompt distillation loss is formulated as
\begin{equation}
\zeta _{\mathrm{PD}}=\sum_{\boldsymbol{x}\in \mathcal{D}}{{\mathrm{KL}\left( p_{\mathrm{soft}}\left(\cdot|\boldsymbol{x} \right) ,p_{\mathrm{hard}}\left( \cdot|\boldsymbol{x} \right)  \right)}},
  \label{eq_PD}
\end{equation}
where $\mathrm{KL}\left( \cdot ,\cdot \right) $ denotes the KL divergence. By minimizing $\zeta _{\mathrm{PD}}$, we can transfer useful knowledge contained in the teacher prompt to the student prompt.

\subsubsection{Mixture-of-Prompts Distillation (MoPD)} 
Usually there may exist significant variations among images within the same dataset \cite{proda}, and hence a single hard prompt is insufficient to accommodate the preferences of all images. To alleviate this issue, it is intuitive to utilize multiple hard prompts to fit such preferences, and a straightforward way is to randomly select hard prompts from a hard prompt pool. However, this method still does not take the individual preferences of each image into consideration and cannot filter out potentially noisy hard prompts from the pool.

To find a better way to distill the knowledge from multiple hard prompts, inspired by mixture-of-experts \cite{moe}, the proposed MoPD method additionally introduces a gating network $G$ to select suitable and instance-specific hard prompts for each image to guide the learning of the soft prompt. For simplicity, the gating network $G$ has only one fully connected layer with a trainable weight matrix $\boldsymbol{W}_g$ and adopts the softmax function as the activation function. The input of the gating network $G$ is the visual embedding $\boldsymbol{f}$ produced by the image encoder. Then the gating network can be formulated as
\begin{equation}
G( \boldsymbol{f} ) = \mathrm{Softmax}( \mathrm{KeepTop} ( \boldsymbol{f}\cdot \boldsymbol{W}_g ,T ) ) ,
  \label{eq_G}
\end{equation}
where $\mathrm{Softmax}(\cdot)$ denotes the softmax function and $\mathrm{KeepTop}(\cdot,\cdot)$ is defined as
{\small
\begin{equation}
\mathrm{KeepTop}(u,T) = 
\begin{cases}
	u_i, &\mathrm{if}\;u_i\;\mathrm{is}\;\mathrm{in}\;\mathrm{top}\;T\;\mathrm{elements}\;\mathrm{of}\;u\\
	-\infty, &\mathrm{otherwise}
\end{cases}.
\label{eq_topk}
\end{equation}
}
\noindent The output of the gating network represents the preference of the image towards all the hard prompts. Since not all the hard prompts are informative, we only choose the top-$T$ hard prompts via the $\mathrm{KeepTop}(\cdot,\cdot)$ operator.

Once the hard prompts are selected, we formulate a mixture-of-prompts distillation loss to distill knowledge from these hard prompts (i.e., teacher prompts) to the soft prompt (i.e., student prompt) as
\begin{equation}
\zeta _{\mathrm{MPD}}=\sum_{t=1}^H{\sum_{\boldsymbol{x}\in \mathcal{D}}{{G\left( \boldsymbol{f} \right) _t \mathrm{KL}\left( p_{\mathrm{soft}}\left( \cdot|\boldsymbol{x} \right) ,p_{\mathrm{tea},t}\left( \cdot|\boldsymbol{x} \right)  \right)}}},
  \label{eq_mpd}
\end{equation}
where $p_{\mathrm{tea},t}\left( \cdot|\boldsymbol{x} \right) $ denotes the prediction probability distribution of the $t$-th teacher prompt for all the classes based on Eq. \eqref{eq_p_hard}, $G( \boldsymbol{f} ) _t$ denotes the weight corresponding to $t$-th teacher prompt, and $H$ denotes the total number of hard prompts in the pool. 

To enhance the gating network's ability to autonomously select the most suitable teacher prompts for downstream tasks and effectively mitigate interference from noisy prompts, we incorporate the prediction probability of teacher prompts into the loss function to guide the learning process of this gating network. Specifically, by encouraging the gating network to assign higher weights to teacher prompts that have higher prediction probabilities for the correct label, the gating network learns to prioritize the most informative and relevant teacher prompts for the downstream task, while suppressing the influence of noisy or less relevant prompts.
Therefore, we propose the mixture-of-prompts selection loss as
\begin{equation}
\zeta _{\mathrm{MPS}}=-\sum_{t=1}^H{\sum_{(\boldsymbol{x},y) \in \mathcal{D}}{{G\left( \boldsymbol{f} \right) _t\ln p_{\mathrm{tea}, t}\left( y|\boldsymbol{x} \right)}}}.
\label{eq_MPS}
\end{equation}

By combining the above considerations together, the overall objective of the MoPD method is formulated as
\begin{equation}
\zeta =\alpha \zeta _{\mathrm{CE}}+\left( 1-\alpha \right) \zeta _{\mathrm{MPD}}+\beta \zeta _{\mathrm{MPS}},
  \label{eq_finalloss}
\end{equation}
where $\alpha$ and $\beta$ are trade-off parameters. Here $\zeta _{\mathrm{CE}}$ and $\zeta _{\mathrm{MPD}}$ are used to learn the soft prompt, while $\zeta _{\mathrm{MPD}}$ and $\zeta _{\mathrm{MPS}}$ help learn the gating network.

Note that the gating network is only used in the training stage. After it is learned in the training process, it will not be used in the inference stage, making the inference process of MoPD identical to other soft prompt learning methods.

\subsection{Discussion}
To see why using multiple hard prompts could be effective, we can give an analysis from the perspective of mixture-of-experts (MoE).

From a technical perspective, the gating mechanism in MoPD is analogous to the MoE paradigm \cite{moe}. In MoE, a gating network dynamically routes inputs to specialized experts based on input features, while in MoPD, the gating network acts as a ``prompt router'' that selects instance-specific hard prompts, which act similarly to experts in MoE, based on image features.

From the theoretical perspective, since the gating mechanism of MoPD is grounded in MoE, the theoretical principles of MoE could be applicable to MoPD. Theorem 4.1 in \cite{theorem_moe} states that ``single expert performs poorly'', while Theorem 4.2 in \cite{theorem_moe} demonstrates that a router can learn cluster-center features, helping divide a complex problem into simpler sub-problems that individual experts can conquer. \cite{theorem_moe} combines Theorem 4.1 and Theorem 4.2 to conclude that ``an MoE provably outperforms a single expert''. Given the connection between MoPD and MoE, those theorems provide additional theoretical support for the proposed MoPD method that could outperform SiPD.

\section{Experiments}
In this section, we evaluate MoPD on \edit{four} settings: (1) base-to-new generalization, (2) few-shot classification, (3) domain generalization, \edit{(4) cross-dataset evaluation}. We also analyze the effect of each module and parameter through ablation study and parameter analysis. We further analyze the impact of noisy prompts.

\paragraph{Datasets} Following \cite{coop, cocoop, prograd, kgcoop}, the base-to-new generalization, \edit{cross-dataset evaluation,} and few-shot image classification are conducted on 11 image classification datasets, including ImageNet \cite{imagenet} and Caltech101 \cite{caltech101} for generic object classification, OxfordPets \cite{oxfordpets}, StanfordCars \cite{stanfordcars}, Flowers102 \cite{oxfordflowers}, Food101 \cite{food}, and FGVCAircraft \cite{FGVCAircraft} for fine-grained visual categorization, SUN397 \cite{sun397} for scene recognition, DTD \cite{dtd} for texture classification, EuroSAT \cite{eurosat} for satellite image classification, and UCF101 \cite{ucf101} for action recognition. Besides, the domain generalization is conducted on ImageNet and its variants, including ImageNetV2 \cite{imagenet_v2}, ImageNet-Sketch \cite{imagenet_sketch}, ImageNet-A \cite{imagenet_a}, and ImageNet-R \cite{imagenet_r}.

\paragraph{Training Details} Our implementation is based on the official implementation of CoOp \cite{coop} and KgCoOp \cite{kgcoop}. All experiments are conducted with ViT-B/16 as the image encoder and Transformer \cite{transformer} as the text encoder.
The gating network is a single-layer fully-connected neural network with the softmax function as the activation function. The length of prompting vectors is fixed to 4, and they are initialized with the template ``a photo of a''. The training schedule and data augmentation settings are identical to those of CoOp and KgCoOp. All experimental results are averaged over three seeds. In Eq. \eqref{eq_finalloss}, $\alpha$ is set to 0.8 by default, except that $\alpha$ is set to 0.5 on Food101, EuroSAT, and UCF101. $\beta$ is set to 0.0005 by default, except that $\beta$ is set to 1.0 on StanfordCars and DTD. The number of selected teacher prompts (i.e., $T$) in the gating network is set to 3 on ImageNet, Caltech101, OxfordPets, StanfordCars, and Flowers102, while for the rest datasets, $T$ is set to 2. The hard prompt pool consisting of task-related hard prompts is simply designed through synonym substitution and sentence transformation on templates of CLIP and CoOp (see Supplementary Material for a full list), and the total number of hard prompts (i.e., $H$) in the pool is fixed to 12. Similar to \cite{cocoop}, we utilize the harmonic mean accuracy
$\mathrm{H}=2/\left( 1/\mathrm{acc}_{\mathrm{Base}} +1/\mathrm{acc}_{\mathrm{New}} \right)$ 
as the metric to quantify the trade-off in generalization, where $\mathrm{acc}_{\mathrm{Base}}$ and $\mathrm{acc}_{\mathrm{New}}$ denote the accuracy of the model on the base and new classes, respectively. A high value of H indicates that the corresponding model performs well on both base and new classes simultaneously.
We use PyTorch to implement all experiments on NVIDIA GeForce RTX 3090.

\paragraph{Baselines} 
We compare with several state-of-the-art soft prompt learning methods, including both \edit{textual prompt (TP) learning methods} (e.g., CoOp \cite{coop}, CoCoOp \cite{cocoop}, ProGrad \cite{prograd}, and KgCoOp \cite{kgcoop}) and \edit{multi-modal prompt (MMP) learning methods (e.g., MaPLe \cite{maple} and MetaPrompt \cite{metaprompt})}. \edit{Hand-crafted prompt (HP) method} (i.e., CLIP \cite{clip}) is also included in comparison to evaluate the generalization performance of hard prompts.

\edit{Since MoPD is optimized using traditional textual prompt learning, it has fewer learnable parameters than multi-modal prompt learning methods (e.g., MaPLe contains 600+ times more learnable parameters than MoPD). Given those fundamental architectural differences, direct comparisons would be methodologically unsuitable. For a fair comparison with MMP methods, we extend the MoPD method into an independent vision-language prompting (IVLP) version named MoPD-MMP, which learns hierarchical prompts for both the text and image encoders separately as \cite{promptsrc}, \cite{ivlp} did.}

\begin{table*}[!htbp]
\renewcommand{\arraystretch}{0.94}
\centering
\caption{Performance of various methods under the base-to-new generalization setting. All methods are trained with 16 instances per base class. `H' denotes the harmonic mean accuracy.}
\vspace{-2.0em}
\subfloat[\textbf{Average over 11 datasets}]{
    \begin{minipage}[t]{0.32\textwidth}
    \setlength{\tabcolsep}{1.5mm}{
    \begin{tabular}{c|l|cc|c}
    \toprule
          & Methods & Base  & New   & H \\
    \midrule
    HP    & CLIP  & 69.34  & 74.22  & 71.70  \\
    \midrule
    \multirow{5}[1]{*}{TP} & CoOp  & 82.64  & 68.00  & 74.61  \\
          & CoCoOp & 80.47  & 71.69  & 75.83  \\
          & ProGrad & \textbf{82.48}  & 70.72  & 76.15  \\
          & KgCoOp & 80.73  & 73.61  & 77.01  \\
          & \cellcolor{Gray}MoPD & \cellcolor{Gray}81.40  & \cellcolor{Gray}\textbf{74.69}  & \cellcolor{Gray}\textbf{77.90}  \\
    \midrule   
    \multirow{3}[1]{*}{\edit{MMP}} & \edit{MaPLe} & \edit{81.95} & \edit{75.14} & \edit{78.40} \\
          & \edit{MetaPrompt} & \edit{82.73} & \edit{75.21} & \edit{78.79} \\
          & \cellcolor{Gray}\edit{MoPD-MMP} & \cellcolor{Gray}\edit{\textbf{83.43}} & \cellcolor{Gray}\edit{\textbf{75.41}} & \cellcolor{Gray}\edit{\textbf{79.22}} \\
    \bottomrule
    \end{tabular}}
    \end{minipage}
}
\hfill
\subfloat[ImageNet]{
    \begin{minipage}[t]{0.32\textwidth}
    \setlength{\tabcolsep}{1.5mm}{
    \begin{tabular}{c|l|cc|c}
    \toprule
          & Methods & Base  & New   & H \\
    \midrule
    HP    & CLIP  & 72.43  & 68.14  & 70.22  \\
    \midrule
    \multirow{5}[1]{*}{TP} & CoOp  & 76.46  & 66.31  & 71.02  \\
          & CoCoOp & 75.98  & \textbf{70.43}  & 73.10  \\
          & ProGrad & \textbf{77.02}  & 66.66  & 71.47  \\
          & KgCoOp & 75.83  & 69.96  & 72.78  \\
          & \cellcolor{Gray}MoPD & \cellcolor{Gray}76.87  & \cellcolor{Gray}\textbf{70.43}  & \cellcolor{Gray}\textbf{73.51}  \\
    \midrule  
    \multirow{3}[1]{*}{\edit{MMP}} & \edit{MaPLe} & \edit{76.66} & \edit{70.54} & \edit{73.47} \\
    & \edit{MetaPrompt} & \edit{77.10} & \edit{\textbf{70.90}} & \edit{\textbf{73.87}} \\
          & \cellcolor{Gray}\edit{MoPD-MMP} & \cellcolor{Gray}\edit{\textbf{77.60}} & \cellcolor{Gray}\edit{69.91} & \cellcolor{Gray}\edit{73.55} \\
    \bottomrule
    \end{tabular}}
    \end{minipage}
 }
\hfill
\subfloat[Caltech101]{
    \begin{minipage}[t]{0.32\textwidth}
    \setlength{\tabcolsep}{1.5mm}{
    \begin{tabular}{c|l|cc|c}
    \toprule
          & Methods & Base  & New   & H \\
    \midrule
    HP    & CLIP  & 96.84  & 94.00  & 95.40  \\
    \midrule
    \multirow{5}[1]{*}{TP} & CoOp  & \textbf{98.11}  & 93.52  & 95.76  \\
          & CoCoOp & 97.96  & 93.81  & 95.84  \\
          & ProGrad & 98.02  & 93.89  & 95.91  \\
          & KgCoOp & 97.72  & 94.39  & 96.03  \\
          & \cellcolor{Gray}MoPD & \cellcolor{Gray}98.07  & \cellcolor{Gray}\textbf{94.90} & \cellcolor{Gray}\textbf{96.46} \\
    \midrule  
    \multirow{3}[1]{*}{\edit{MMP}} & \edit{MaPLe} & \edit{97.74} & \edit{\textbf{94.36}} & \edit{96.02} \\
          & \edit{MetaPrompt} & \edit{98.27} & \edit{94.10} & \edit{96.14} \\
          & \cellcolor{Gray}\edit{MoPD-MMP} & \cellcolor{Gray}\edit{\textbf{98.36}} & \cellcolor{Gray}\edit{94.18} & \cellcolor{Gray}\edit{\textbf{96.22}} \\
    \bottomrule
    \end{tabular}}
    \end{minipage}
 }
\\
\subfloat[OxfordPets]{
    \begin{minipage}[t]{0.32\textwidth}
    \setlength{\tabcolsep}{1.5mm}{
    \begin{tabular}{c|l|cc|c}
    \toprule
          & Methods & Base  & New   & H \\
    \midrule
    HP    & CLIP  & 91.17  & 97.26  & 94.12  \\
    \midrule
    \multirow{5}[1]{*}{TP} & CoOp  & 94.24  & 96.66  & 95.43  \\
          & CoCoOp & \textbf{95.20}  & 97.67  & \textbf{96.42}  \\
          & ProGrad & 95.07  & 97.63  & 96.33  \\
          & KgCoOp & 94.65  & \textbf{97.76}  & 96.18  \\
          & \cellcolor{Gray}MoPD & \cellcolor{Gray}95.17  & \cellcolor{Gray}97.47  & \cellcolor{Gray}96.31  \\
    \midrule  
    \multirow{3}[1]{*}{\edit{MMP}} & \edit{MaPLe} & \edit{95.43} & \edit{97.76} & \edit{96.58} \\
    & \edit{MetaPrompt} & \edit{\textbf{95.57}} & \edit{96.87} & \edit{96.22} \\
          & \cellcolor{Gray}\edit{MoPD-MMP} & \cellcolor{Gray}\edit{95.43} & \cellcolor{Gray}\edit{\textbf{97.78}} & \cellcolor{Gray}\edit{\textbf{96.59}} \\
    \bottomrule
    \end{tabular}}
    \end{minipage}
}
\hfill
\subfloat[StanfordCars]{
    \begin{minipage}[t]{0.32\textwidth}
    \setlength{\tabcolsep}{1.5mm}{
    \begin{tabular}{c|l|cc|c}
    \toprule
          & Methods & Base  & New   & H \\
    \midrule
    HP    & CLIP  & 63.37  & 74.89  & 68.65  \\
    \midrule
    \multirow{5}[1]{*}{TP} & CoOp  & 76.20  & 69.14  & 72.50  \\
          & CoCoOp & 70.49  & 73.59  & 72.01  \\
          & ProGrad & \textbf{77.68}  & 68.63  & 72.88  \\
          & KgCoOp & 71.76  & \textbf{75.04} & 73.36  \\
          & \cellcolor{Gray}MoPD & \cellcolor{Gray}75.63  & \cellcolor{Gray}73.10  & \cellcolor{Gray}\textbf{74.34}  \\
    \midrule  
    \multirow{3}[1]{*}{\edit{MMP}} & \edit{MaPLe} & \edit{72.94} & \edit{74.00} & \edit{73.47} \\
    & \edit{MetaPrompt} & \edit{75.47} & \edit{\textbf{74.67}} & \edit{75.07} \\
          & \cellcolor{Gray}\edit{MoPD-MMP} & \cellcolor{Gray}\edit{\textbf{79.75}} & \cellcolor{Gray}\edit{73.06} & \cellcolor{Gray}\edit{\textbf{76.26}} \\
    \bottomrule
    \end{tabular}}
    \end{minipage}
}
\hfill
\subfloat[Flowers102]{
    \begin{minipage}[t]{0.32\textwidth}
    \setlength{\tabcolsep}{1.5mm}{
    \begin{tabular}{c|l|cc|c}
    \toprule
          & Methods & Base  & New   & H \\
    \midrule
    HP    & CLIP  & 72.08  & 77.80 & 74.83  \\
    \midrule
    \multirow{5}[1]{*}{TP} & CoOp  & \textbf{97.63} & 69.55  & 81.23  \\
          & CoCoOp & 94.87  & 71.75  & 81.71  \\
          & ProGrad & 95.54  & 71.87  & 82.03  \\
          & KgCoOp & 95.00  & \textbf{74.73}  & 83.65  \\
          & \cellcolor{Gray}MoPD & \cellcolor{Gray}95.80  & \cellcolor{Gray}74.43  & \cellcolor{Gray}\textbf{83.77}  \\
    \midrule  
    \multirow{3}[1]{*}{\edit{MMP}} & \edit{MaPLe} & \edit{95.92} & \edit{72.46} & \edit{82.56} \\
    & \edit{MetaPrompt} & \edit{97.20} & \edit{74.13} & \edit{84.11} \\
          & \cellcolor{Gray}\edit{MoPD-MMP} & \cellcolor{Gray}\edit{\textbf{97.28}} & \cellcolor{Gray}\edit{\textbf{75.18}} & \cellcolor{Gray}\edit{\textbf{84.81}} \\
    \bottomrule
    \end{tabular}}
    \end{minipage}
}
\hfill
\\
\subfloat[Food101]{
    \begin{minipage}[t]{0.32\textwidth}
    \setlength{\tabcolsep}{1.5mm}{
    \begin{tabular}{c|l|cc|c}
    \toprule
          & Methods & Base  & New   & H \\
    \midrule
    HP    & CLIP  & 90.10  & 91.22  & 90.66  \\
    \midrule
    \multirow{5}[1]{*}{TP} & CoOp  & 89.44  & 87.50  & 88.46  \\
          & CoCoOp & 90.70  & 91.29  & 90.99  \\
          & ProGrad & 90.37  & 89.59  & 89.98  \\
          & KgCoOp & 90.50  & \textbf{91.70}  & 91.10  \\
          & \cellcolor{Gray}MoPD & \cellcolor{Gray}\textbf{90.77}  & \cellcolor{Gray}91.67  & \cellcolor{Gray}\textbf{91.22}  \\
    \midrule  
    \multirow{3}[1]{*}{\edit{MMP}} & \edit{MaPLe} & \edit{90.71} & \edit{\textbf{92.05}} & \edit{\textbf{91.38}} \\
    & \edit{MetaPrompt} & \edit{90.70} & \edit{91.67} & \edit{91.18} \\
          & \cellcolor{Gray}\edit{MoPD-MMP} & \cellcolor{Gray}\edit{\textbf{90.85}} & \cellcolor{Gray}\edit{91.75} & \cellcolor{Gray}\edit{91.30} \\
    \bottomrule
    \end{tabular}}
    \end{minipage}
}
\hfill
\subfloat[FGVCAircraft]{
    \begin{minipage}[t]{0.32\textwidth}
    \setlength{\tabcolsep}{1.5mm}{
    \begin{tabular}{c|l|cc|c}
    \toprule
          & Methods & Base  & New   & H \\
    \midrule
    HP    & CLIP  & 27.19  & 36.29  & 31.09  \\
    \midrule
    \multirow{5}[1]{*}{TP} & CoOp  & 39.24  & 30.49  & 34.32  \\
          & CoCoOp & 33.41  & 23.71  & 27.74  \\
          & ProGrad & \textbf{40.54}  & 27.57  & 32.82  \\
          & KgCoOp & 36.21  & 33.55  & 34.83  \\
          & \cellcolor{Gray}MoPD & \cellcolor{Gray}37.43  & \cellcolor{Gray}\textbf{35.13}  & \cellcolor{Gray}\textbf{36.24}  \\
    \midrule  
    \multirow{3}[1]{*}{\edit{MMP}} & \edit{MaPLe} & \edit{37.44} & \edit{35.61} & \edit{36.50} \\
    & \edit{MetaPrompt} & \edit{38.50} & \edit{\textbf{36.60}} & \edit{37.53} \\
          & \cellcolor{Gray}\edit{MoPD-MMP} & \cellcolor{Gray}\edit{\textbf{41.18}} & \cellcolor{Gray}\edit{35.55} & \cellcolor{Gray}\edit{\textbf{38.16}} \\
    \bottomrule
    \end{tabular}}
    \end{minipage}
}
\hfill
\subfloat[SUN397]{
    \begin{minipage}[t]{0.32\textwidth}
    \setlength{\tabcolsep}{1.5mm}{
    \begin{tabular}{c|l|cc|c}
    \toprule
          & Methods & Base  & New   & H \\
    \midrule
    HP    & CLIP  & 69.36  & 75.35  & 72.23  \\
    \midrule
    \multirow{5}[1]{*}{TP} & CoOp  & 80.85  & 68.34  & 74.07  \\
          & CoCoOp & 79.74  & 76.86  & 78.27  \\
          & ProGrad & \textbf{81.26}  & 74.17  & 77.55  \\
          & KgCoOp & 80.29  & \textbf{76.53}  & 78.36  \\
          & \cellcolor{Gray}MoPD & \cellcolor{Gray}81.13  & \cellcolor{Gray}76.47  & \cellcolor{Gray}\textbf{78.73}  \\
    \midrule  
    \multirow{3}[1]{*}{\edit{MMP}} & \edit{MaPLe} & \edit{80.82} & \edit{\textbf{78.70}} & \edit{79.75} \\
    & \edit{MetaPrompt} & \edit{81.80} & \edit{78.50} & \edit{80.12} \\
          & \cellcolor{Gray}\edit{MoPD-MMP} & \cellcolor{Gray}\edit{\textbf{82.20}} & \cellcolor{Gray}\edit{78.54} & \cellcolor{Gray}\edit{\textbf{80.33}} \\
    \bottomrule
    \end{tabular}}
    \end{minipage}
}
\\
\subfloat[DTD]{
    \begin{minipage}[t]{0.32\textwidth}
    \setlength{\tabcolsep}{1.5mm}{
    \begin{tabular}{c|l|cc|c}
    \toprule
          & Methods & Base  & New   & H \\
    \midrule
    HP    & CLIP  & 53.24  & 59.90  & 56.37  \\
    \midrule
    \multirow{5}[1]{*}{TP} & CoOp  & \textbf{80.17}  & 47.54  & 59.69  \\
          & CoCoOp & 77.01  & 56.00  & 64.85  \\
          & ProGrad & 77.35  & 52.35  & 62.44  \\
          & KgCoOp & 77.55  & 54.99  & 64.35  \\
          & \cellcolor{Gray}MoPD & \cellcolor{Gray}77.27  & \cellcolor{Gray}\textbf{57.47}  & \cellcolor{Gray}\textbf{65.92}  \\
    \midrule  
    \multirow{3}[1]{*}{\edit{MMP}} & \edit{MaPLe} & \edit{80.36} & \edit{\textbf{59.18}} & \edit{68.16} \\
    & \edit{MetaPrompt} & \edit{\textbf{82.20}} & \edit{58.97} & \edit{\textbf{68.67}} \\
          & \cellcolor{Gray}\edit{MoPD-MMP} & \cellcolor{Gray}\edit{81.40} & \cellcolor{Gray}\edit{57.65} & \cellcolor{Gray}\edit{67.50} \\
    \bottomrule
    \end{tabular}}
    \end{minipage}
}
\hfill
\subfloat[EuroSAT]{
    \begin{minipage}[t]{0.32\textwidth}
    \setlength{\tabcolsep}{1.5mm}{
    \begin{tabular}{c|l|cc|c}
    \toprule
          & Methods & Base  & New   & H \\
    \midrule
    HP    & CLIP  & 56.48  & 64.05  & 60.03  \\
    \midrule
    \multirow{5}[1]{*}{TP} & CoOp  & \textbf{91.54} & 54.44  & 68.28  \\
          & CoCoOp & 87.49  & 60.04  & 71.21  \\
          & ProGrad & 90.11  & 60.89  & 72.67  \\
          & KgCoOp & 85.64  & 64.34  & 73.48  \\
          & \cellcolor{Gray}MoPD & \cellcolor{Gray}85.77  & \cellcolor{Gray}\textbf{72.50}  & \cellcolor{Gray}\textbf{78.58}  \\
    \midrule  
    \multirow{3}[1]{*}{\edit{MMP}} & \edit{MaPLe} & \edit{\textbf{90.47}} & \edit{73.23} & \edit{80.94} \\
    & \edit{MetaPrompt} & \edit{88.77} & \edit{72.50} & \edit{79.81} \\
          & \cellcolor{Gray}\edit{MoPD-MMP} & \cellcolor{Gray}\edit{86.79} & \cellcolor{Gray}\edit{\textbf{78.20}} & \cellcolor{Gray}\edit{\textbf{82.27}} \\
    \bottomrule
    \end{tabular}}
    \end{minipage}
}
\hfill
\subfloat[UCF101]{
    \begin{minipage}[t]{0.32\textwidth}
    \setlength{\tabcolsep}{1.5mm}{
    \begin{tabular}{c|l|cc|c}
    \toprule
          & Methods & Base  & New   & H \\
    \midrule
    HP    & CLIP  & 70.53  & 77.50  & 73.85  \\
    \midrule
    \multirow{5}[1]{*}{TP} & CoOp  & \textbf{85.14}  & 64.47  & 73.38  \\
          & CoCoOp & 82.33  & 73.45  & 77.64  \\
          & ProGrad & 84.33  & 74.94  & 79.36  \\
          & KgCoOp & 82.89  & 76.67  & 79.66  \\
          & \cellcolor{Gray}MoPD & \cellcolor{Gray}81.53  & \cellcolor{Gray}\textbf{78.07}  & \cellcolor{Gray}\textbf{79.76}  \\
    \midrule  
    \multirow{3}[1]{*}{\edit{MMP}} & \edit{MaPLe} & \edit{83.00} & \edit{\textbf{78.66}} & \edit{80.77} \\
    & \edit{MetaPrompt} & \edit{84.47} & \edit{78.40} & \edit{81.32} \\
          & \cellcolor{Gray}\edit{MoPD-MMP} & \cellcolor{Gray}\edit{\textbf{86.85}} & \cellcolor{Gray}\edit{77.68} & \cellcolor{Gray}\edit{\textbf{82.01}} \\
    \bottomrule
    \end{tabular}}
    \end{minipage}
}
\label{table_base2new}
\end{table*}

\subsection{Results on Base-to-New Generalization}
We follow the base-to-new generalization setting of \mbox{CoCoOp}, where all classes are split into disjointed base and new groups and all the methods in comparison are trained on the base classes and evaluated on base and new classes. The experimental results shown in Table \ref{table_base2new} reveal the following observations.

\begin{enumerate}
\item{In terms of the average harmonic mean accuracy, MoPD outperforms all \edit{TP and HP} baseline methods, showcasing its excellent generalization capability on the base-to-new setting. Specifically, MoPD attains the highest harmonic mean accuracy in 10 out of 11 datasets.}
\item{In terms of the average performance on the new classes, MoPD surpasses all \edit{TP and HP} baseline methods. MoPD exhibits a significant performance improvement over soft prompt learning methods and outperforms CLIP on 7 out of 11 datasets. This demonstrates that the selected teacher prompts can effectively guide the soft prompt in acquiring useful knowledge and enhancing its generalization capability.}
\item{In terms of the average performance on the base classes, CoOp and ProGrad perform better among all methods \edit{(except MetaPrompt and MoPD-MMP)} since they may overfit the base classes, leading to poor performance on new classes. MoPD performs slightly inferior to CoOp and ProGrad on the base classes, but substantially outperforms them on the new classes, while MoPD outperforms CoCoOp and KgCoOp on both the base and new classes.}
\item{\edit{Compared with the HP and TP baseline methods, the MMP methods can achieve a better trade-off on the performance between base and new classes. In other words, the MMP methods outperform the HP and TP methods in terms of the average harmonic mean accuracy. Compared with all baseline methods including the MMP methods (i.e., MaPLe and MetaPrompt), MoPD-MMP achieves the best average performance on base classes, new classes, and harmonic mean accuracy, which demonstrates the effectiveness of the proposed MoPD method and its good compatibility with multi-modal prompt learning.}}
\end{enumerate}

\subsection{Results on Few-shot Classification}
\begin{figure*}[!tbph]
  \subfloat{
    \includegraphics[width=0.29\linewidth]{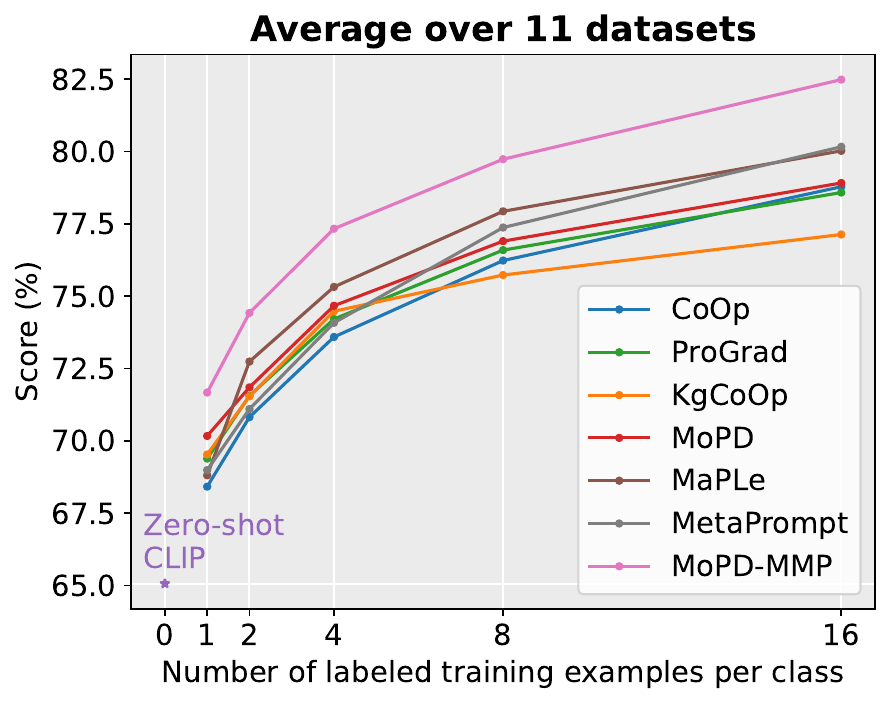}
}
  \hfill
  \subfloat{
    \includegraphics[width=0.29\linewidth]{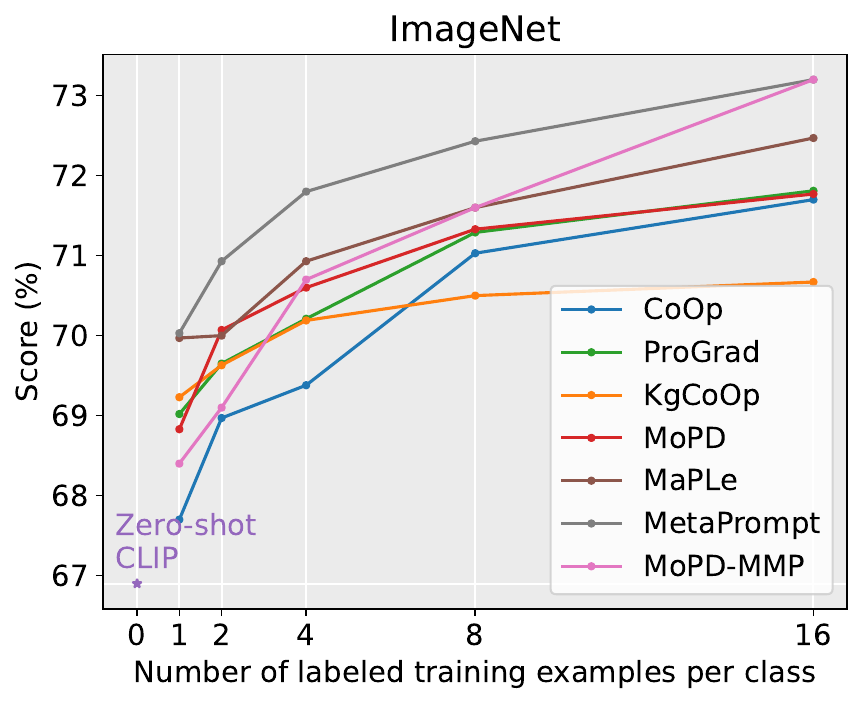}
}
  \hfill
  \subfloat{
    \includegraphics[width=0.29\linewidth]{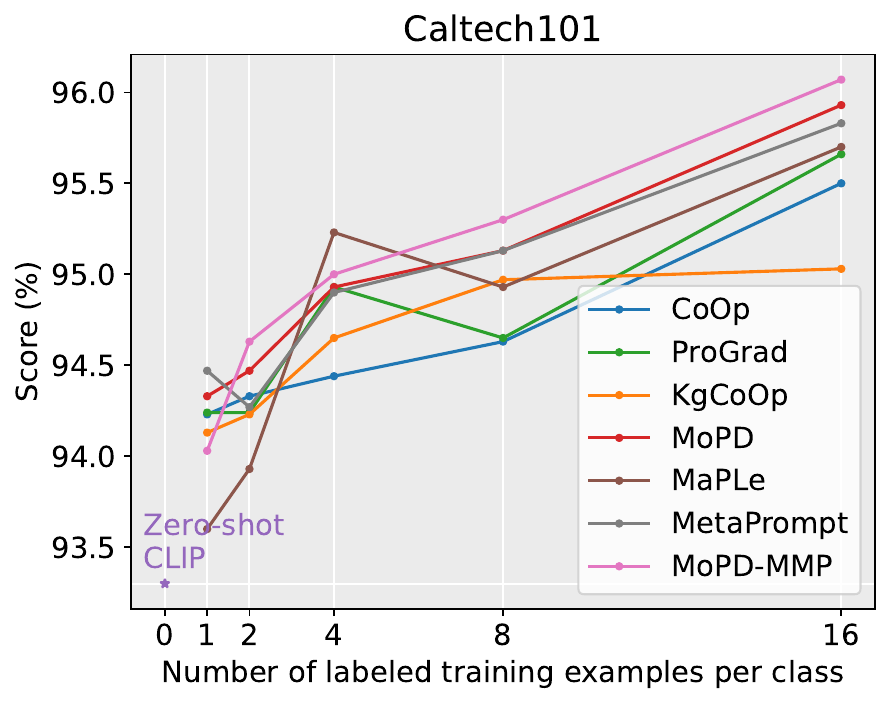}
}
    
  \subfloat{
    \includegraphics[width=0.29\linewidth]{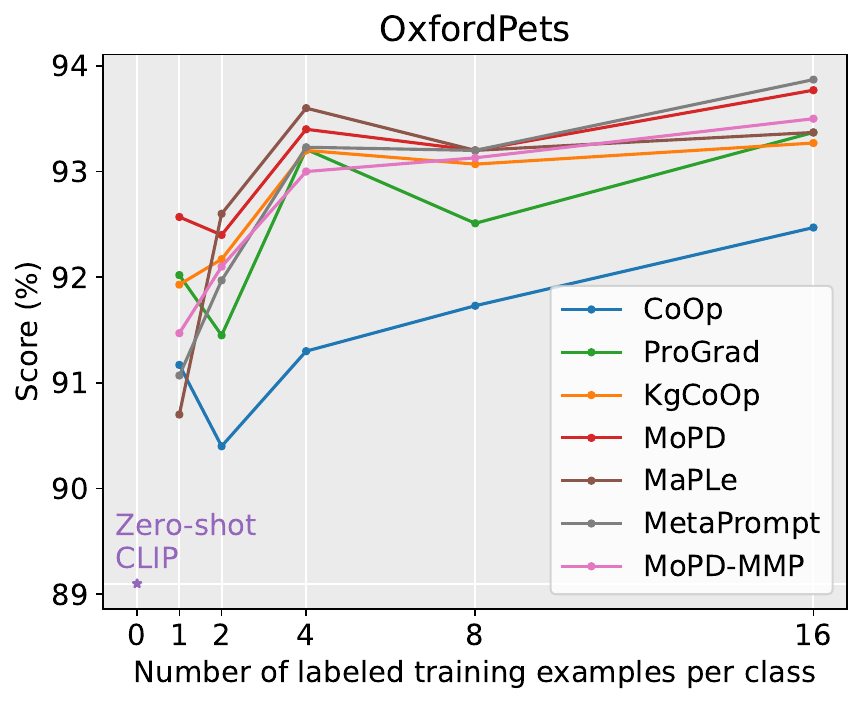}
}
\hfill
    \subfloat{
    \includegraphics[width=0.29\linewidth]{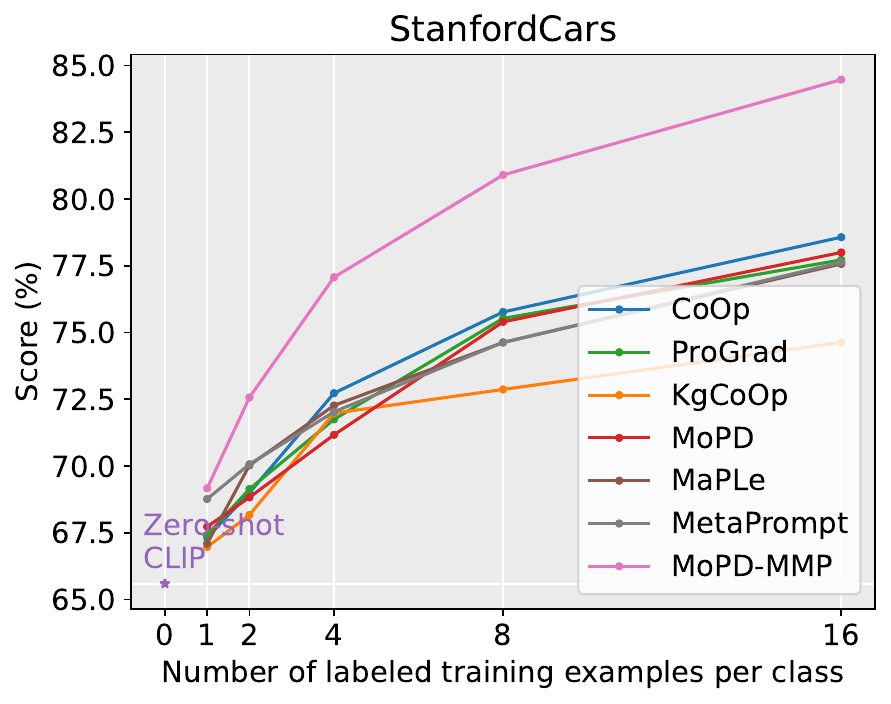}
}
  \hfill
      \subfloat{
    \includegraphics[width=0.29\linewidth]{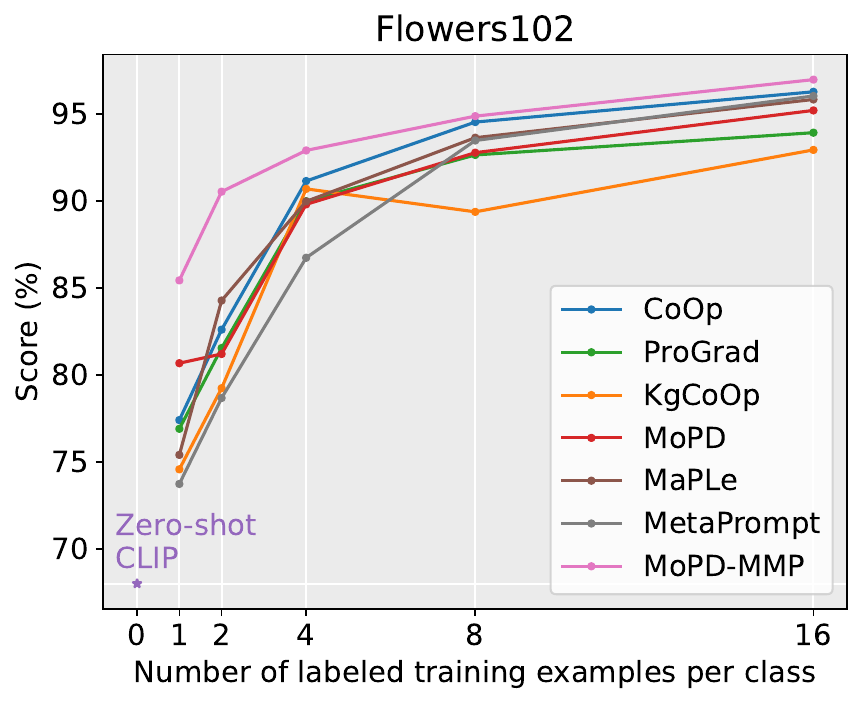}
}

    \subfloat{
    \includegraphics[width=0.29\linewidth]{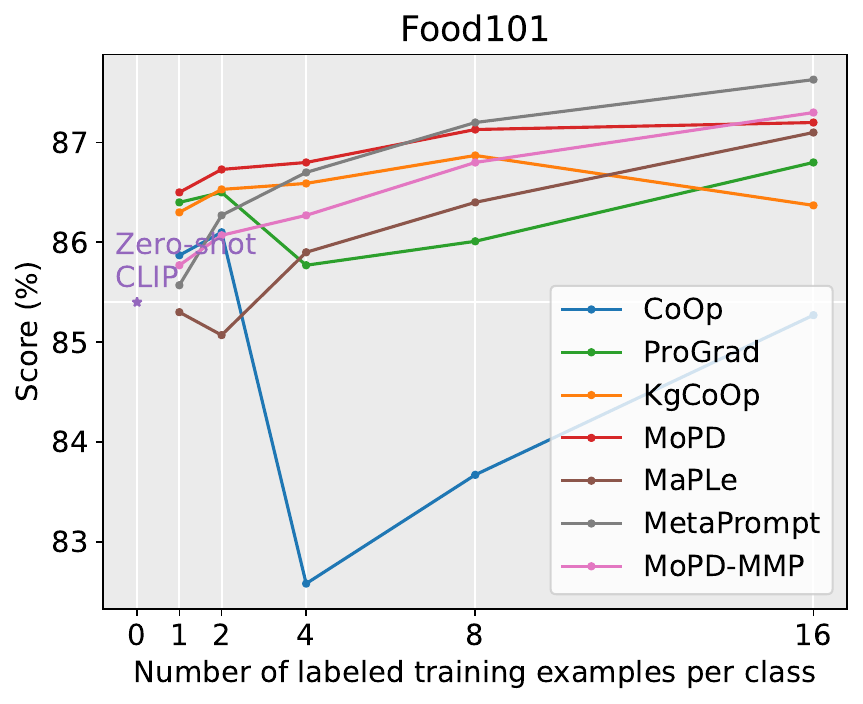}
}
\hfill
    \subfloat{
    \includegraphics[width=0.29\linewidth]{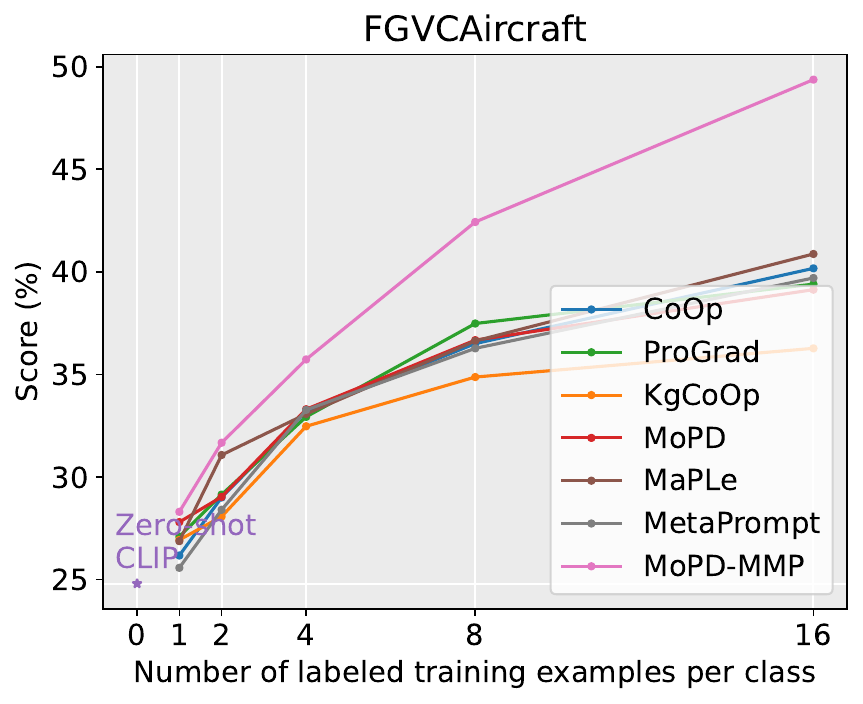}
}
  \hfill
      \subfloat{
    \includegraphics[width=0.29\linewidth]{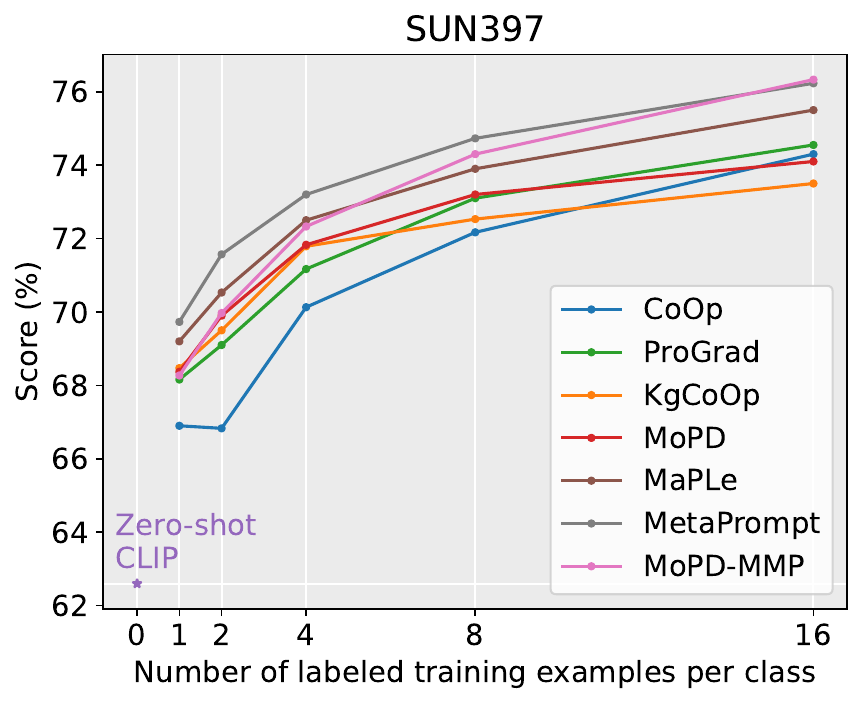}
}

      \subfloat{
    \includegraphics[width=0.29\linewidth]{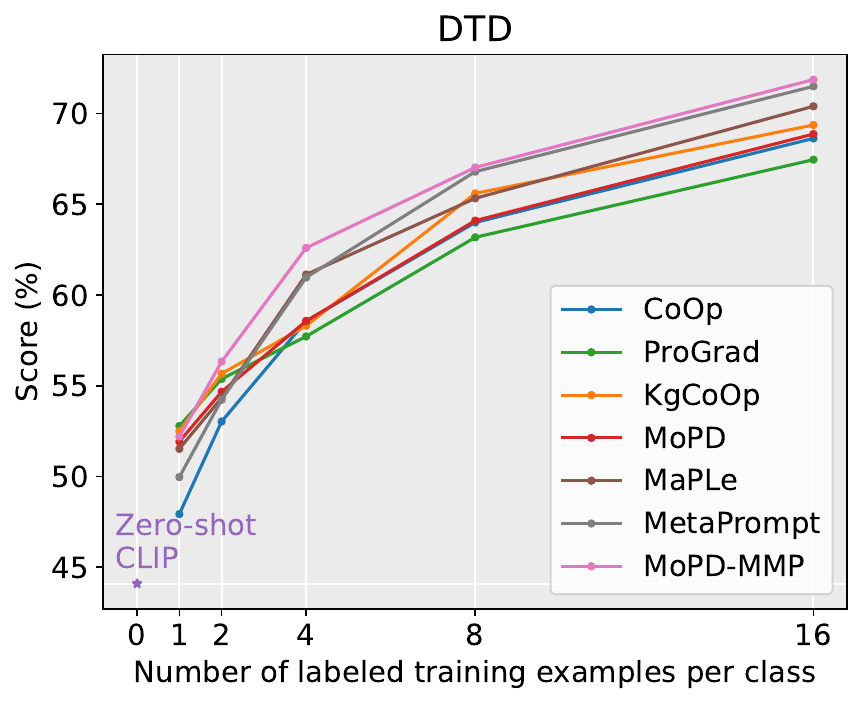}
}
\hfill
    \subfloat{
    \includegraphics[width=0.29\linewidth]{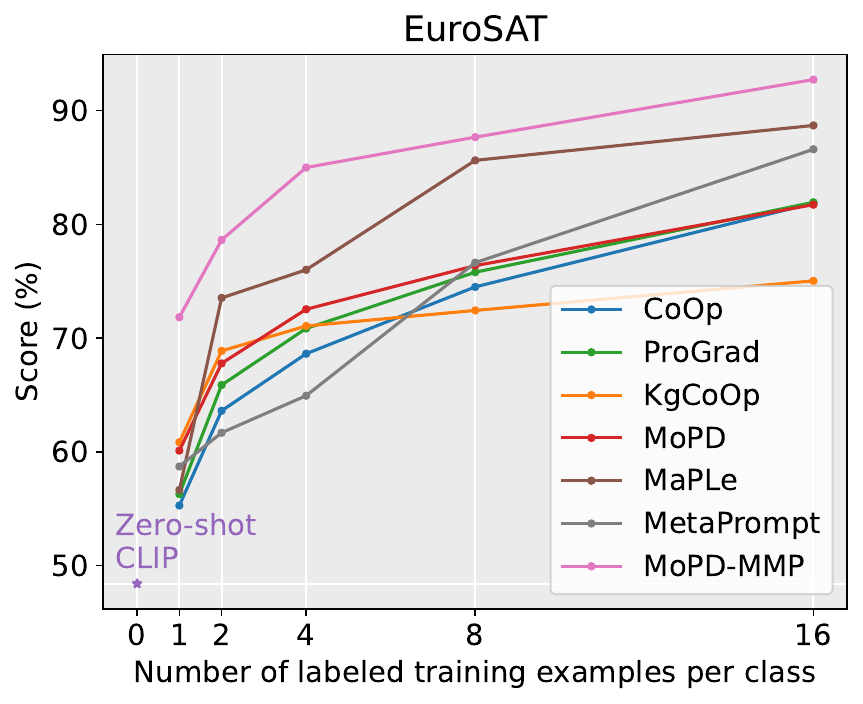}
}
  \hfill
      \subfloat{
    \includegraphics[width=0.29\linewidth]{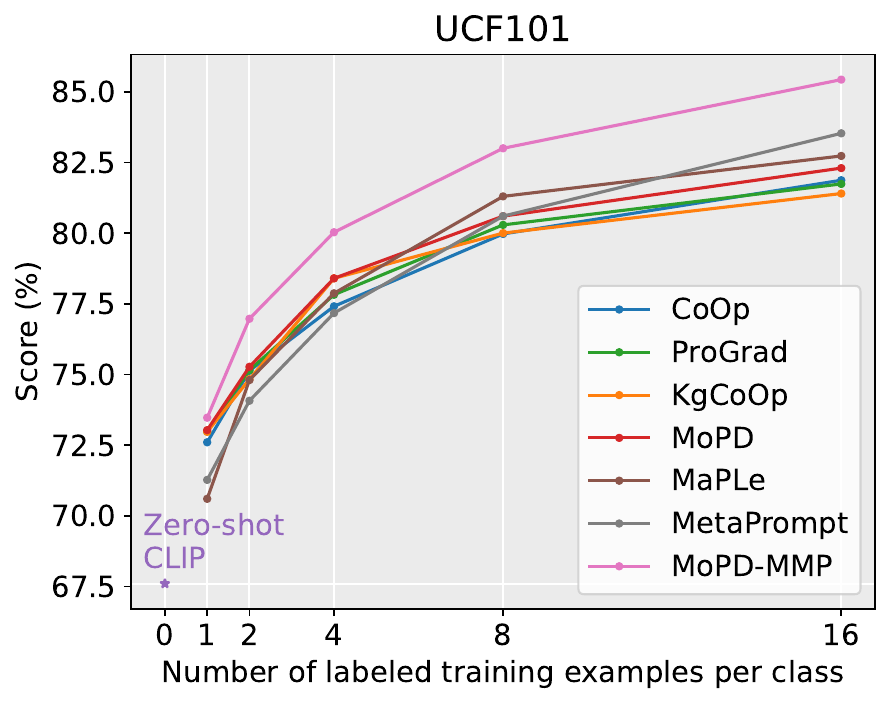}
}
  \caption{\edit{Performance of various methods under the few-shot classification setting on the 11 datasets.}}
  \label{fig_few_shot}
\end{figure*}

In the few-shot classification setting, each method is trained on all the classes of a dataset with few-shot labeled samples and evaluated on the rest data of this dataset. Following \cite{coop}, we perform few-shot classification with different numbers of shots, specifically 1, 2, 4, 8, and 16 shots. The performance of all the methods on the 11 datasets is shown in Fig. \ref{fig_few_shot}. 
According to the results, MoPD achieves the best average performance among all \edit{TP and HP} baseline methods. \edit{While MaPLe and MetaPrompt outperform the TP methods in terms of the average performance under 8 and 16 shots, their effectiveness diminishes under 1, 2, and 4 shots, achieving only comparable performance to TP methods. This limitation likely arises from insufficient parameter optimization under extreme data scarcity. In contrast, MoPD-MMP exhibits better average performance than all baseline methods under all shots.}
Those results further demonstrate the good generalization ability of the proposed MoPD and \edit{MoPD-MMP} methods even when the samples of downstream tasks are limited.

\subsection{Results on Domain Generalization}
Generalisation to out-of-distribution data is a necessary capability for VLMs. The domain generalization setting is employed to evaluate the generalization ability of a model trained on a source domain to a target domain that shares the same classes as the source domain but exhibits a different data distribution. As with all baseline methods (except zero-shot CLIP), we perform soft prompt learning on the source domain (i.e., ImageNet) and evaluate the model on four different target domains (i.e., ImageNetV2, ImageNet-Sketch, Imagenet-A, and ImageNet-R).

According to results shown in Table \ref{table_domaingeneralization}, we can observe that MoPD \edit{and MoPD-MMP} outperform all baseline methods except ProGrad \edit{and MetaPrompt} in the source domain.
In terms of the average performance on all four target domains,
\edit{MoPD and MoPD-MMP outperform all baseline methods and especially MoPD-MMP possesses the highest overall average accuracy of 60.63\%}, which confirms that the soft prompts learned in MoPD are more domain-generalizable.

\begin{table}[!htbp]
  \centering
  \caption{Performance of various methods under the domain generalization setting. All methods use 16-shot samples from each of the 1000 classes on ImageNet except CLIP. The best and second-best results are marked in bold and underlined.}
    \fontsize{9pt}{11pt}\selectfont
    \setlength{\tabcolsep}{1.5mm}{
    \begin{tabular}{c|c|cccc|c}
    \toprule
          & Source & \multicolumn{5}{c}{Target} \\
\cmidrule{2-7}          & ImageNet & -V2   & -Sketch & -A    & -R    & Avg. \\
    \midrule
    CLIP  & 66.73  & 60.83  & 46.15  & 47.77  & 73.96  & 57.17  \\
    CoOp  & 71.51  & 64.20  & 47.99  & 49.71  & 75.21  & 59.28  \\
    CoCoOp & 71.02  & 64.07  & 48.75  & 50.63  & 76.18  & 59.90  \\
    ProGrad & \underline{72.24} & 64.73  & 47.61  & 49.39  & 74.58  & 59.07  \\
    KgCoOp & 71.20  & 64.10  & 48.97  & 50.69  & 76.70  & 60.11  \\
    \rowcolor{Gray} MoPD  & 71.77  & 64.70  & 49.10  & \underline{50.77}  & 76.60  & \underline{60.29}  \\
    \edit{MaPLe} & \edit{70.72} & \edit{64.07} & \edit{49.15} & \edit{\textbf{50.90}} & \edit{76.98} & \edit{60.28} \\
    \edit{MetaPrompt} & \edit{\textbf{72.60}} & \edit{\textbf{65.50}} & \edit{\textbf{49.30}} & \edit{49.03} & \edit{\textbf{77.20}} & \edit{60.28} \\
    \rowcolor{Gray} \edit{MoPD-MMP} & \edit{71.93} & \edit{\underline{65.40}} & \edit{\textbf{49.30}} & \edit{50.70} & \edit{\underline{77.10}} & \edit{\textbf{60.63}} \\
    \bottomrule
    \end{tabular}}
  \label{table_domaingeneralization}%
\end{table}%

\subsection{\edit{Results on Cross-Dataset Evaluation}}
\edit{In the cross-dataset evaluation setting that evaluates whether prompts learned from a source domain can generalize to unseen target domains, prompts are trained using 16-shot samples from each of the 1000 classes on ImageNet and then evaluated on the other 10 datasets. According to results shown in Table \ref{tab_cross}, the proposed MoPD and MoPD-MMP methods perform comparably to the SOTA method MaPLe while using significantly fewer tunable parameters (8K vs. 52K vs. 3.55M), and demonstrate competitive performance compared to other baseline methods.}

\begin{table*}[htbp]
  \centering
  \caption{\edit{Performance of various methods under the cross-dataset evaluation setting. All methods use 16-shot samples from each of the 1000 classes on ImageNet except CLIP. The best results are marked in bold. The numbers in the parentheses following each method name indicate the number of tunable parameters.}}
      \resizebox{\linewidth}{!}{
    \setlength{\tabcolsep}{1mm}{
    \begin{tabular}{c|c|cccccccccc|c}
    \toprule
          & Source & \multicolumn{11}{c}{Target} \\
\cmidrule{2-13}          & ImageNet & Caltech101 & OxfordPets & StanfordCars & Flowers102 & Food101 & FGVCAircraft & SUN397 & DTD   & EuroSAT & UCF101 & Avg. \\
    \midrule
    CoOp (2K) & 71.51  & 93.70  & 89.14  & 64.51  & 68.71  & 85.30  & 18.47  & 64.15  & 41.92  & 46.39  & 66.55  & 63.88  \\
    CoCoOp (35K) & 71.02  & \textbf{94.43} & 90.14  & 65.32  & 71.88  & 86.06  & 22.94  & \textbf{67.36} & 45.73  & 45.37  & 68.21  & 65.74  \\
    ProGrad (2K) & 72.24 & 91.52  & 89.64  & 62.39  & 67.87  & 85.40  & 20.61  & 62.47  & 39.42  & 43.46  & 64.29  & 62.71  \\
    KgCoOp (2K) & 70.66  & 93.92  & 89.83  & 65.41  & 70.01  & 86.36  & 22.51  & 66.16  & 46.35  & 46.04  & 68.50  & 65.51  \\
    \rowcolor{Gray} MoPD (8K)  & 71.77  & 94.20  & 90.50  & \textbf{66.30} & 71.60  & \textbf{86.80} & 23.90  & 67.00  & \textbf{47.10} & 43.80  & \textbf{69.60} & 66.08  \\
    \edit{MaPLe (3.55M)} & \edit{70.72} & \edit{93.53} & \edit{90.49} & \edit{65.57} & \edit{\textbf{72.23}} & \edit{86.20} & \edit{\textbf{24.74}} & \edit{67.01} & \edit{46.49} & \edit{48.06} & \edit{68.69} & \edit{\textbf{66.30}} \\
    \edit{MetaPrompt (61K)} & \edit{\textbf{72.60}} & \edit{93.83} & \edit{89.87} & \edit{65.13} & \edit{70.83} & \edit{85.63} & \edit{24.50} & \edit{67.10} & \edit{46.73} & \edit{47.17} & \edit{66.47} & \edit{65.73} \\
    \rowcolor{Gray} \edit{MoPD-MMP (52K)} & \edit{71.93} & \edit{94.30} & \edit{\textbf{90.70}} & \edit{65.70} & \edit{71.10} & \edit{86.30} & \edit{23.40} & \edit{66.90} & \edit{46.40} & \edit{\textbf{48.30}} & \edit{68.90} & \edit{66.20} \\
    \bottomrule
    \end{tabular}}}
  \label{tab_cross}%
\end{table*}%

\subsection{Ablation Study}
We conduct the ablation study to demonstrate the effect of mixture-of-prompts distillation loss $\zeta _{\mathrm{MPD}}$, mixture-of-prompts selection loss $\zeta _{\mathrm{MPS}}$, and the selection of gating network under the base-to-new generalization setting. The results are listed in Table \ref{ablation}, where the baseline is CoOp, SiPD built on CoOp is formulated in Section \ref{sec_MoPD}, MoPD-R denotes a variant of MoPD by randomly selecting teacher prompts instead of using the gating network, and MoPD (w/o $\zeta _{\mathrm{MPS}}$) denotes MoPD without minimizing $\zeta _{\mathrm{MPS}}$.

\noindent\textbf{Effect of mixture-of-prompts distillation loss.} Firstly, to demonstrate the effect of prompt distillation technique, we compare CoOp with SiPD (incorporating prompt distillation loss $\zeta _{\mathrm{PD}}$). The results reveal that, in comparison to CoOp, SiPD trades a slight decrease in the average accuracy on base classes (i.e., -1.53$\%$) for a significant improvement in accuracy on new classes (i.e., +5.35$\%$) and harmonic mean accuracy (i.e., + 2.43$\%$) over all 11 datasets. This illustrates that teacher prompts can effectively guide the soft prompt to learn useful knowledge and enhance its generalization capability through prompt distillation. Furthermore, compared with SiPD, MoPD (incorporating mixture-of-prompts distillation loss $\zeta _{\mathrm{MPD}}$) performs better on base classes (i.e., +0.29$\%$ improvement), new classes (i.e., +1.34$\%$ improvement), and harmonic mean accuracy (i.e., +0.86$\%$ improvement), highlighting the advantages of minimizing the mixture-of-prompts distillation loss, which helps soft prompts to learn from multiple hard prompts rather than a single hard prompt.

\noindent\textbf{Effect of mixture-of-prompts selection loss.} In comparison to MoPD (w/o $\zeta _{\mathrm{MPS}}$), MoPD
minimizes $\zeta _{\mathrm{MPS}}$, enabling the gating network to select informative and relevant teacher prompts for the downstream task. Specifically, MoPD outperforms MoPD (w/o $\zeta _{\mathrm{MPS}}$) on new classes (i.e., +0.5$\%$ improvement) and harmonic mean accuracy (i.e., +0.21$\%$ improvement) over all 11 datasets, which demonstrates the effectiveness of the mixture-of-prompts selection loss.

\noindent\textbf{Effect of the gating network.} In comparison to MoPD-R, MoPD is able to select instance-specific teacher prompts for the downstream task through the trainable gating network instead of random sampling. MoPD outperforms MoPD-R on new classes (i.e., +1.17$\%$ improvement) and harmonic mean accuracy (i.e., +0.64$\%$ improvement) over all 11 datasets without degrading the accuracy on base classes, and its average performance on new classes even beats CLIP with hard prompts, demonstrating the necessity of employing the gating network to select teacher prompts.

\begin{table}[!htbp]
  \caption{Ablation study for the mixture-of-prompts distillation loss $\zeta _{\mathrm{MPD}}$, mixture-of-prompts selection loss $\zeta _{\mathrm{MPS}}$, and gating network.}
  \centering
  \fontsize{9pt}{11pt}\selectfont
  \setlength{\tabcolsep}{1mm}{
    \begin{tabular}{lcc|cccc}
    \toprule
    Dataset & Set   & Baseline & SiPD  & MoPD-R & MoPD  & MoPD \\
          &       & (CoOp) &       &        & (w/o $\zeta _{\mathrm{MPS}}$) &  \\
    \midrule
    \multirow{3}{*}{Average} & Base  & \textbf{82.64} & 81.11  & 81.39  & 81.54  & 81.40  \\
          & New   & 68.00  & 73.35  & 73.52  & 74.19  & \textbf{74.69} \\
          & H     & 74.61  & 77.04  & 77.26  & 77.69  & \textbf{77.90} \\
    \bottomrule
    \end{tabular}}
  \label{ablation}%
\end{table}%

\noindent \edit{\textbf{Effect of knowledge transfer methods.} We conduct ablation studies on knowledge transfer methods. We evaluated four distinct methods: 1) KL divergence-based knowledge distillation used in MoPD, which minimizes the KL divergence between prediction probability distributions of hard and soft prompts; 2) MMD-based knowledge distillation, which minimizes the Maximum Mean Discrepancy \cite{mmd} between prediction probability distributions of hard and soft prompts; 3) Cosine similarity-based knowledge transfer \cite{kgcoop}, which maximizes cosine similarity between textual embeddings of hard and soft prompts; and 4) $\ell_1$ loss-based knowledge transfer \cite{promptsrc}, which minimizes the $\ell_1$ loss between textual embeddings of hard and soft prompts. According to results shown in Table \ref{tab_knowledge_transfer}, we can see that the proposed MoPD method that is with KL divergence-based knowledge distillation outperforms alternative knowledge transfer methods in terms of average harmonic mean accuracy, demonstrating that the proposed knowledge distillation method in MoPD is effective.}

\begin{table}[htbp]
  \centering
  \caption{\edit{Effect of different knowledge transfer methods on the average performance of all 11 datasets under the base-to-new generalization setting.}}
    \begin{tabular}{l|ccc}
    \toprule
    Method & Base  & New   & H \\
    \midrule
    MoPD-$\ell_1$ & 81.74  & 73.16  & 77.21  \\
    MoPD-cos & 80.24  & 74.46  & 77.24  \\
    MoPD-MMD & 80.63  & 73.18  & 76.72  \\
    \rowcolor{Gray} MoPD (ours) & 81.40  & 74.69  & \textbf{77.90} \\
    \bottomrule
    \end{tabular}%
  \label{tab_knowledge_transfer}%
\end{table}%

\noindent\edit{\textbf{Computational Cost.} The computational cost of MoPD is shown in Table \ref{tab_cost}. We can observe that MoPD introduces negligibly computational overhead compared to SiPD, while MoPD achieves performance gains over SiPD. Those results demonstrate that the proposed MoPD is time-efficient.}

\begin{table}[htbp]
  \centering
  \caption{\edit{Computational cost of MoPD, where the model is trained on the base classes of ImageNet. $H$ denotes the pool's total number of hard prompts, `ms' denotes the millisecond per image, and `H (Avg.)' denotes the average harmonic mean accuracy of 11 datasets.}}
    \setlength{\tabcolsep}{1mm}{
    \begin{tabular}{l|cccc|c}
    \toprule
    Method & \multicolumn{1}{l}{Learnable para.} & Train.time & Infer.time & Memory & H (Avg.) \\
    \midrule
    SiPD  & 4K    & 31min36s & 2.33ms & 9044MB & 77.04 \\
    MoPD  & (4+$H$)K & +15s & +0ms & +5.8MB & 77.90 \\
    \bottomrule
    \end{tabular}}
  \label{tab_cost}%
\end{table}%

\begin{figure*}[tb]
  \subfloat[]
    {\includegraphics[width=0.24\linewidth]{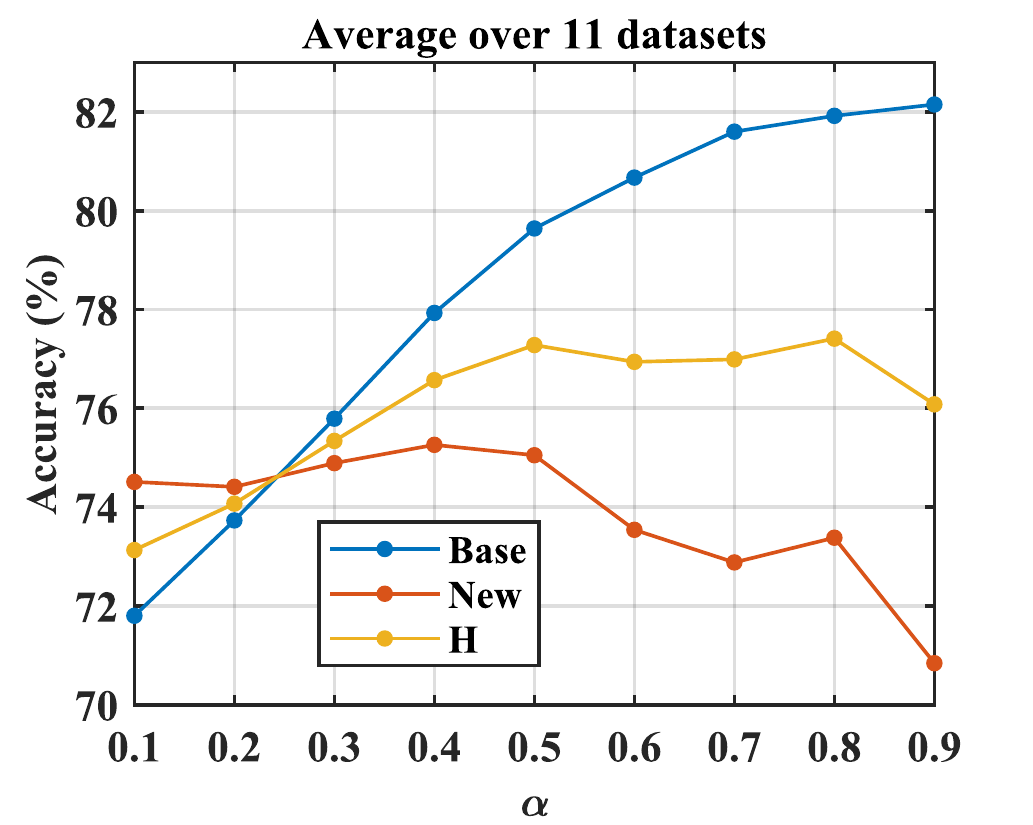}
    \label{alpha}
 }\hfill
  \subfloat[]
    {\includegraphics[width=0.24\linewidth]{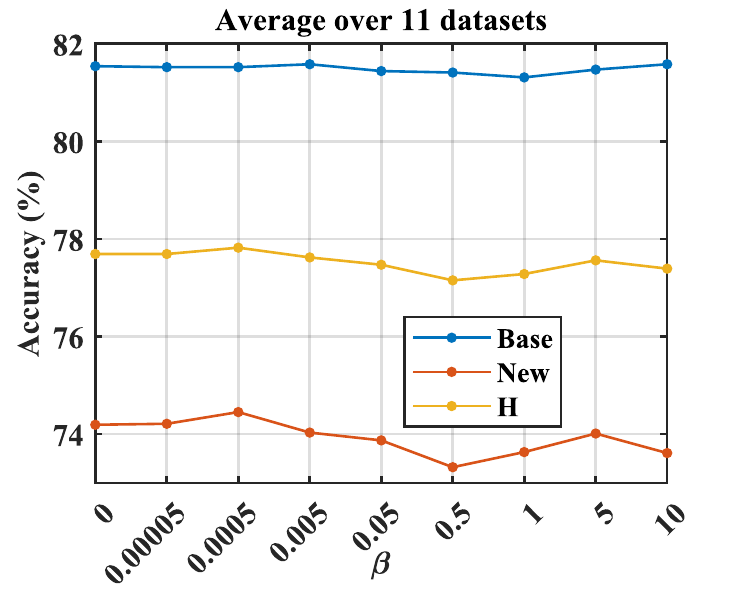}
    \label{beta}}
  \hfill
  \subfloat[]
    {\includegraphics[width=0.24\linewidth]{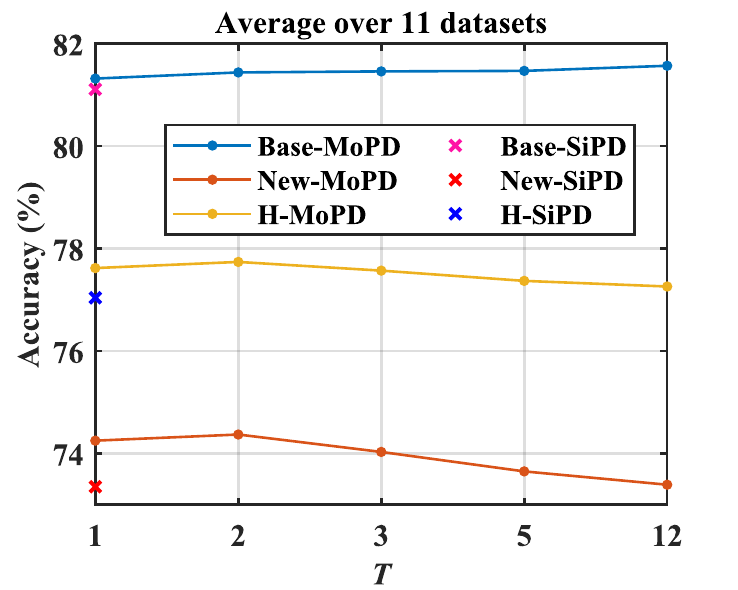}
    \label{T}}
    \hfill
  \subfloat[]
    {\includegraphics[width=0.24\linewidth]{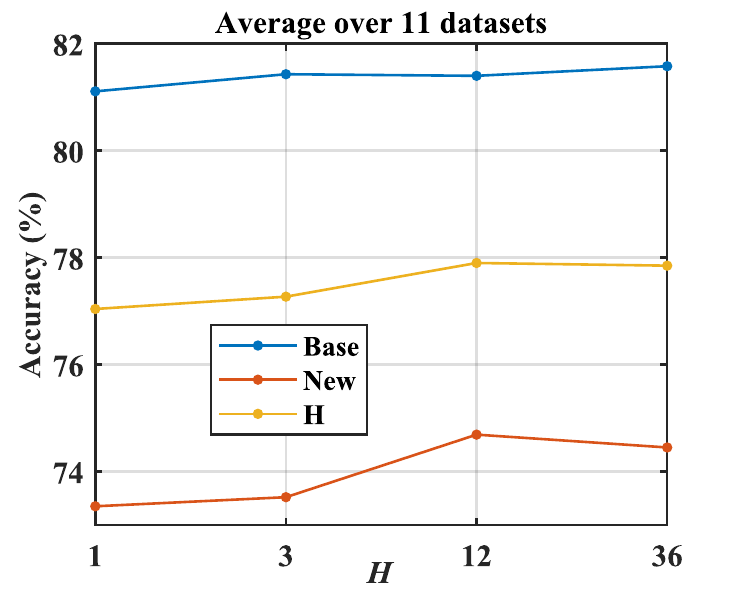}
    \label{H}}
  \caption{\edit{The average performance of all datasets achieved by MoPD for hyperparameters (i.e., $\alpha$, $\beta$, $T$, and $H$) under the base-to-new generalization setting.}}
  \label{fig_parameter}
\end{figure*}

\subsection{Parameter Analysis}
In this section, we analyze the effect of several hyperparameters (i.e., $\alpha$, $\beta$, $T$, and $H$) in MoPD. The average performance achieved by MoPD on all datasets under the base-to-new generalization setting is shown in Fig. \ref{fig_parameter}.

{\bf For the trade-off parameter $\alpha$}, according to Fig. \ref{alpha}, we can see that as $\alpha$ increases, the accuracy on base classes increases, while that on new classes increases and then decreases. In other words, increasing $\alpha$ will increase the weight of $\zeta _{\mathrm{CE}}$, allowing the soft prompt to fit better on base classes, but will deprive its generality on new classes. On the other hand, decreasing $\alpha$ to increase the weight of $\zeta _{\mathrm{MPD}}$ will enable the soft prompt to generalize better on new classes. 
This suggests that it is necessary to choose a medium value for $\alpha$, which is usually set to 0.5 or 0.8 to achieve a better harmonic mean accuracy in our experiments.

{\bf For the trade-off parameter $\beta$}, Fig. \ref{beta} illustrates that $\beta$ within some range (e.g., 0.00005 to 10) has little impact on base classes. In terms of the performance on new classes and their harmonic mean accuracy, the performance shows a trend of initially increasing and then decreasing within some range of $\beta$ (e.g., 0.00005 to 0.5). Those results suggest that $\beta$ could take values around 0.0005.

{\bf For the number of selected teacher prompts $T$}, it is evident from Fig. \ref{T} that as $T$ increases, accuracy on base classes increases slightly, and the accuracy on new classes increases first and then decreases. A large $T$ will weaken the effect of each teacher prompt, leading to inferior performance. When $T=2$, the average H over all datasets is the highest. It is important to note that when $T=1$, the teacher prompt for the same image may differ across different training iterations. Consequently, the image also benefits from guidance provided by multiple teacher prompts. This is a key factor contributing to MoPD's strong performance, where MoPD surpasses SiPD by a considerable margin even when $T=1$.

{\bf For the parameter $H$}, we expand the range of the number of hard prompts in pool from 1 to 36 and as shown in Fig. \ref{H}, the performance of MoPD generally improves as the number of prompts increases from 1 to 12. However, further augmentation of the prompt quantity yields no significant performance enhancement. This observation suggests that continually increasing the number of prompts does not confer substantial benefits.

\subsection{Further Analysis}
\noindent\textbf{Robustness of MoPD.} To investigate the robustness of MoPD to noisy prompts, we add noisy prompts 
into the hard prompt pool and compare MoPD with MoPD-R under the base-to-new generalization setting on the UCF101 and DTD datasets, where noisy prompts are randomly generated based on grammatically plausible sentence templates with incoherent words generated by ChatGPT \cite{chatgpt,gpt4}, such as ``In the whispering forest, a [CLASS] dreams of stars'' (more examples can be found in Supplementary Material). 

According to the results shown in Table \ref{tab_noisy_propmt}, when the hard prompt pool is full of noisy prompts (i.e., 24N and 12N), the performance of both MoPD and MoPD-R is clearly worse than MoPD and MoPD-R with 12T.
Nevertheless, even in situations where the hard prompt pool is exclusively composed of noisy prompts, MoPD is capable of discerning relatively superior teacher prompts, resulting in higher harmonic mean accuracy compared to MoPD-R.
When the hard prompt pool is a mixture of noisy prompts and task-related prompts, the higher the proportion of noisy prompts in the prompt pool, the worse the performance (in terms of base, new, and harmonic mean accuracy) of MoPD-R.
However, the performance of MoPD with 12T shows almost no difference when compared to MoPD with 12T+12N and MoPD with 12T+24N. This is because the gating network can help select suitable teacher prompts. Those experimental results demonstrate the robustness of MoPD against noisy prompts.

Furthermore, we investigate the impact of load balancing \cite{moe} on MoPD in the Supplementary Material. The results indicate that the performance comparison between the balanced and unbalanced versions is negligible. Consequently, in our experiments, we present results for the basic MoPD implementation without load balancing. Additional details and comprehensive analyses of this aspect can be found in the Supplementary Material.

\begin{figure*}[!tbph]
  \subfloat[CLIP]{
    \includegraphics[width=0.29\linewidth]{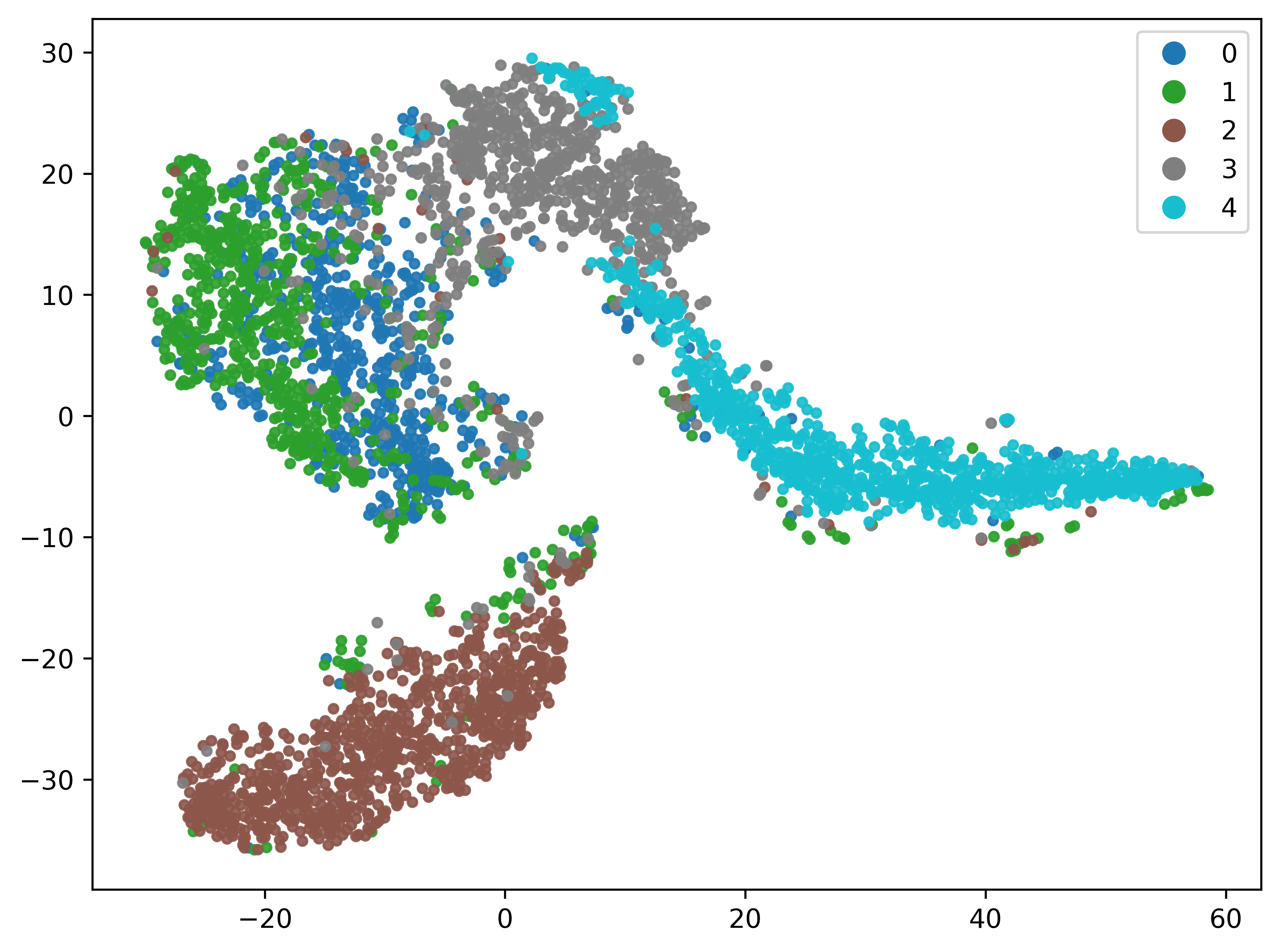}
}
  \hfill
  \subfloat[CoOp]{
    \includegraphics[width=0.29\linewidth]{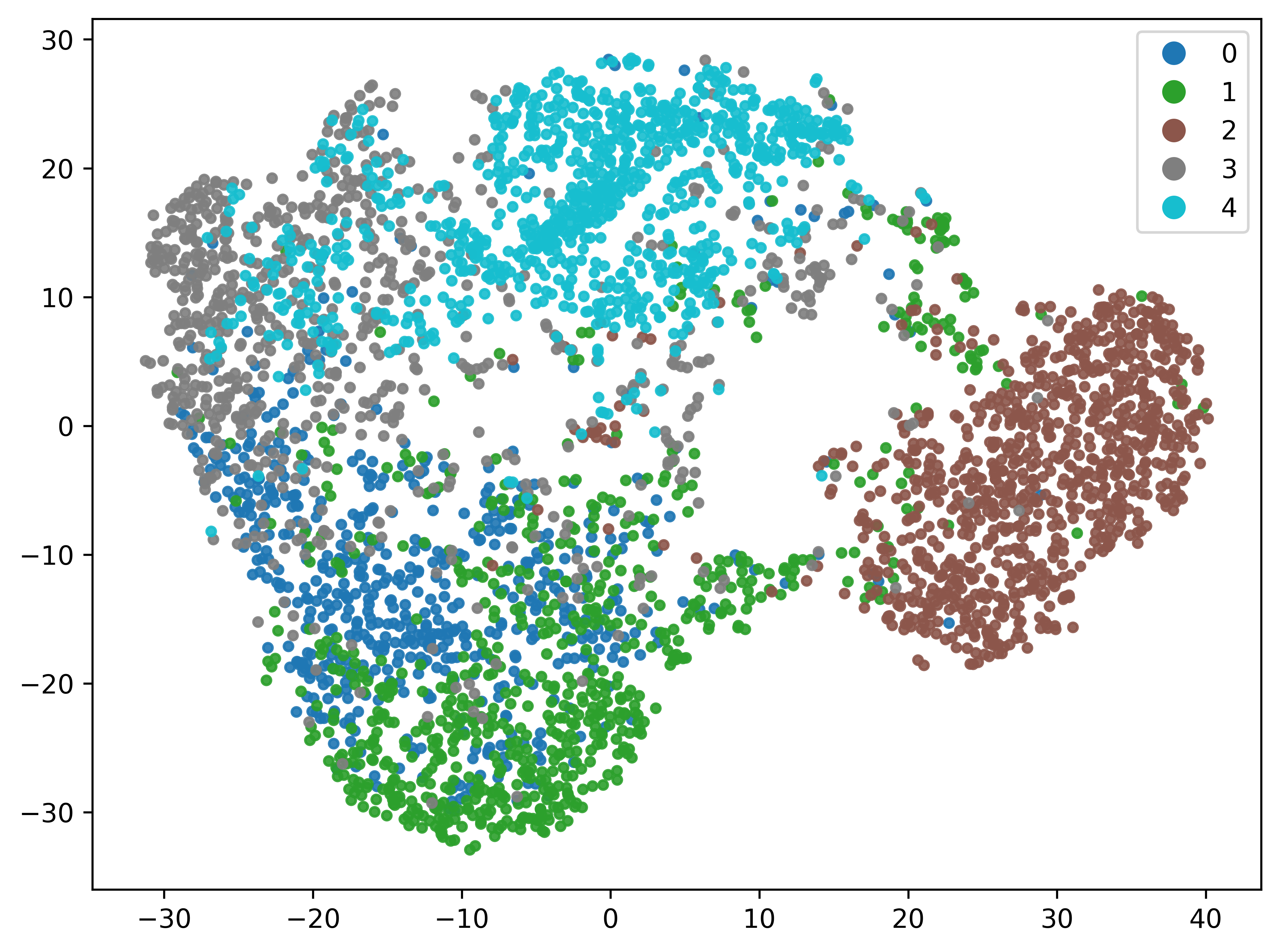}
}
  \hfill
  \subfloat[CoCoOp]{
    \includegraphics[width=0.29\linewidth]{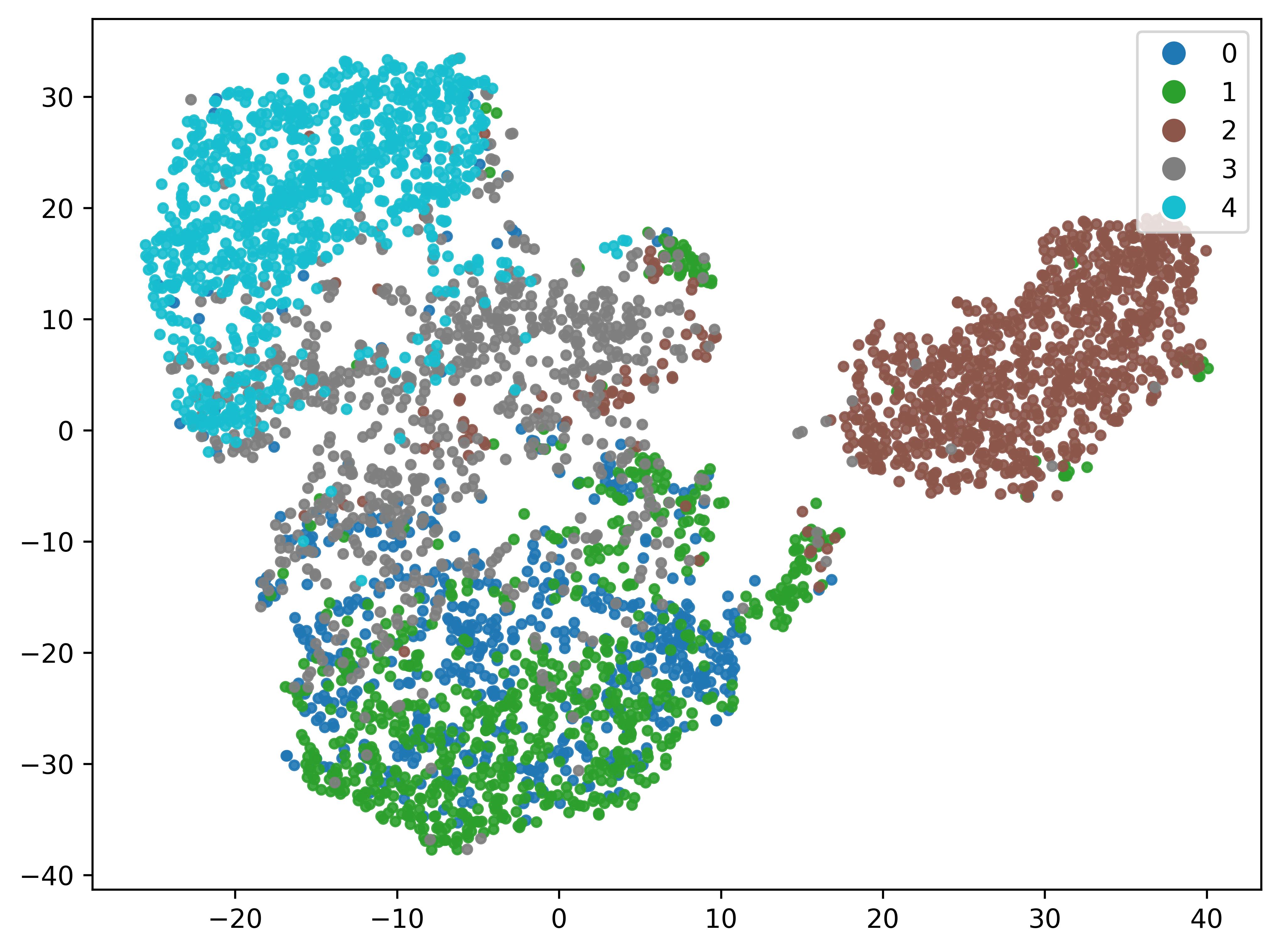}
}

  \subfloat[ProGrad]{
    \includegraphics[width=0.29\linewidth]{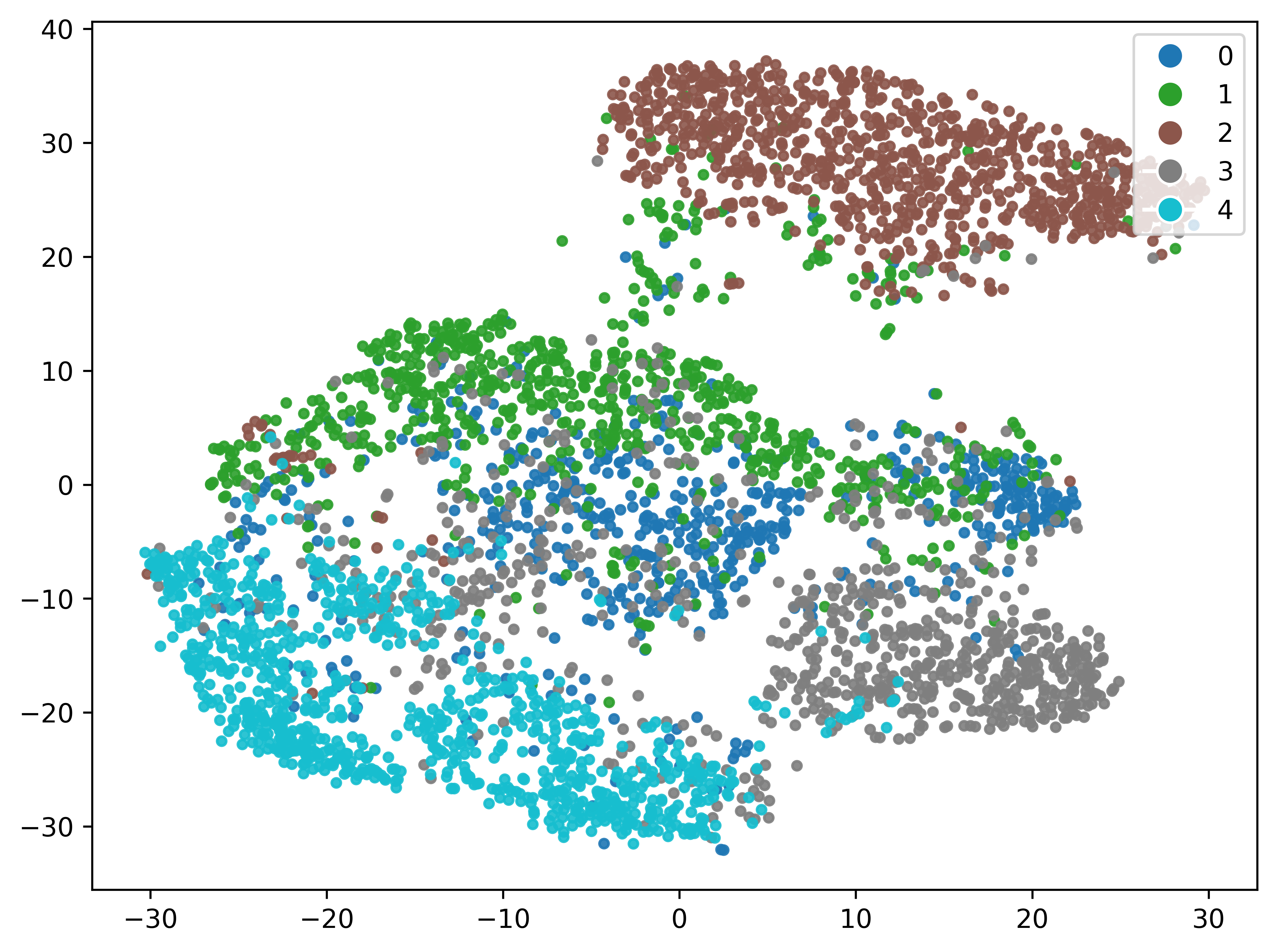}
}
\hfill
    \subfloat[KgCoOp]{
    \includegraphics[width=0.29\linewidth]{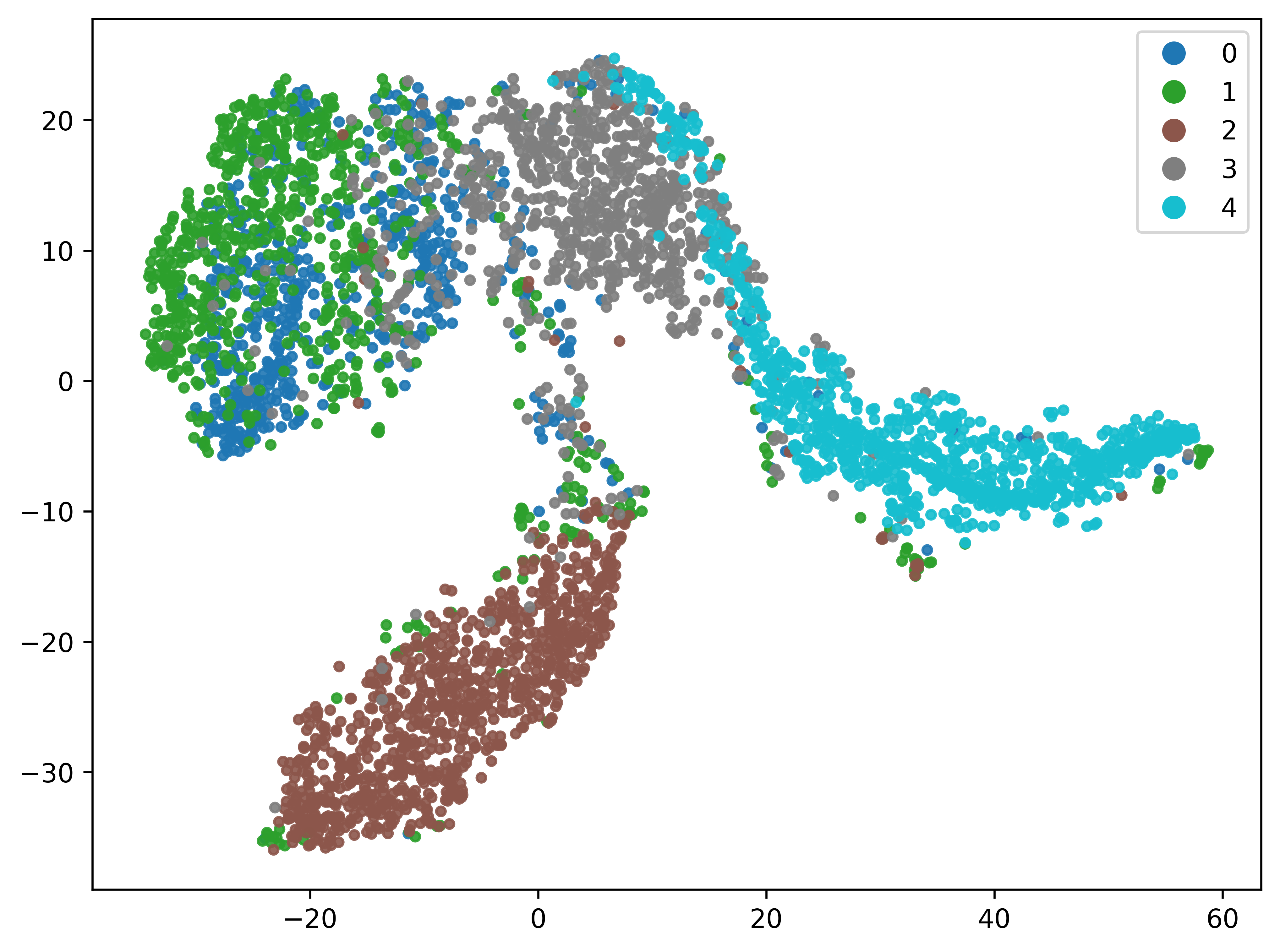}
}
\hfill
      \subfloat[MoPD]{
    \includegraphics[width=0.29\linewidth]{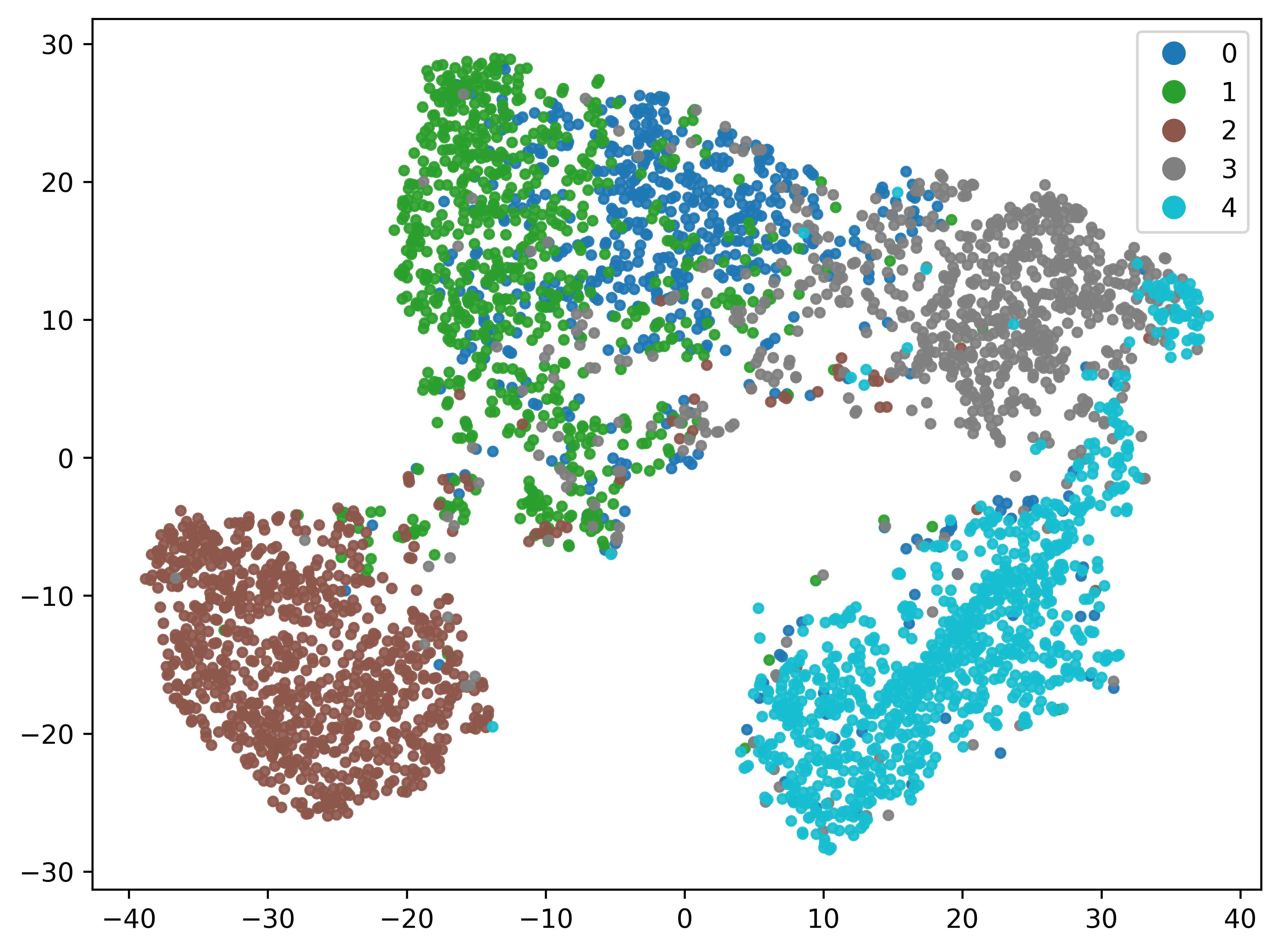}
}
  \caption{\edit{Visualization of different prediction distributions via t-SNE.}}
  \label{fig_tsne}
\end{figure*}

\noindent \edit{\textbf{Visualization.} We use t-SNE \cite{tsne} to analyze prediction distributions on new classes of the EuroSAT dataset under the 16-shot base-class training setting. As shown in Fig. \ref{fig_tsne}, we compare CLIP, CoOp, CoCoOp, ProGrad, KgCoOp, and MoPD, where the numbers 0 to 4 represent the categories of \textit{Pasture Land}, \textit{Permanent Crop Land}, \textit{Residential Buildings}, \textit{River}, and \textit{Sea or Lake}, respectively.
The visualization shows that KgCoOp's prediction manifold closely aligns with CLIP's zero-shot distribution. CoOp, CoCoOp, and ProGrad exhibit larger intra-class variance and ambiguous inter-class boundaries. Notably, MoPD achieves better inter-class separation, particularly for similar category pairs like \textit{Pasture Land} (dark blue) and \textit{Permanent Crop Land} (green), \textit{River} (grey) and \textit{Sea or Lake} (light blue). Those visualization results indicate the enhanced discriminative capability of the proposed MoPD method. In summary, those visualizations corroborate the good generalization of MoPD on new classes.}

\noindent\edit{\textbf{Analysis on Failure Case.} We observe that MoPD underperforms CoCoOp and ProGrad on the OxfordPets dataset under the base-to-new generalization setting. This can be attributed to small category shift between the downstream task and CLIP's pretraining data, as noted in \cite{dept}. Consequently, MoPD extracts limited generalization benefits from hard prompts, leading to inferior results for new classes and harmonic mean accuracy in this specific dataset.}

\begin{table}[tb!]
\caption{Robustness of MoPD to noisy prompts under the base-to-new generalization setting on the UCF101 and DTD datasets. `N' denotes noisy prompts, `T' denotes task-related prompts, and `+' denotes the mixture of the two types of prompts.}
  \centering
  \fontsize{9pt}{11pt}\selectfont
  \setlength{\tabcolsep}{1mm}{
    \begin{tabular}{c|c|ccccc}
    \toprule
    \multirow{8}[8]{*}{UCF101} & MoPD  & 24N   & 12N   & 12T+24N & 12T+12N & 12T \\
\cmidrule{2-7}          & Base  & \textbf{79.03} & \textbf{79.47} & \textbf{81.57} & \textbf{81.60} & 81.53  \\
          & New   & \textbf{74.77} & \textbf{74.30} & \textbf{77.93} & \textbf{78.03} & \textbf{78.07} \\
          & H     & \textbf{76.84} & \textbf{76.80} & \textbf{79.71} & \textbf{79.78} & \textbf{79.76} \\
\cmidrule{2-7}          & MoPD-R & 24N   & 12N   & 12T+24N & 12T+12N & 12T \\
\cmidrule{2-7}          & Base  & 77.93  & 77.77  & 79.40  & 80.33  & \textbf{81.60} \\
          & New   & 73.97  & 74.10  & 75.30  & 77.50  & 77.70  \\
          & H     & 75.90  & 75.89  & 77.30  & 78.89  & 79.60  \\
    \midrule
    \multirow{8}[8]{*}{DTD} & MoPD  & 24N   & 12N   & 12T+24N & 12T+12N & 12T \\
\cmidrule{2-7}          & Base  & 78.20  & \textbf{78.33} & 77.53  & 77.63  & 77.27  \\
          & New   & \textbf{55.63} & \textbf{55.60} & \textbf{57.50} & \textbf{57.63} & \textbf{57.47} \\
          & H     & \textbf{65.01} & \textbf{65.04} & \textbf{66.03} & \textbf{66.15} & \textbf{65.92} \\
\cmidrule{2-7}          & MoPD-R & 24N   & 12N   & 12T+24N & 12T+12N & 12T \\
\cmidrule{2-7}          & Base  & \textbf{78.67} & 78.07  & \textbf{78.73} & \textbf{78.50} & \textbf{78.27} \\
          & New   & 53.80  & 54.60  & 54.03  & 54.80  & 55.57  \\
          & H     & 63.90  & 64.26  & 64.08  & 64.54  & \textbf{64.99} \\
    \bottomrule
    \end{tabular}}
  \label{tab_noisy_propmt}%
\end{table}%

\section{Conclusion}
In this paper, we propose a soft prompt learning method for VLMs named Mixture-of-Prompts Distillation (MoPD) to enhance the generability of soft prompts on unseen classes. MoPD leverages a gating network to select suitable instance-specific teacher prompts for prompt distillation, guiding soft prompts to acquire useful knowledge. Extensive experiments conducted on 11 benchmark datasets demonstrate the effectiveness of MoPD. Notably, MoPD exhibits superior generalization performance on unseen classes, surpassing even CLIP with hard prompts.

\section*{Acknowledgments}

This work is supported by National Key R\&D Program of China (No. 2022ZD0160300) and NSFC grant (No. 62136005).


\bibliographystyle{IEEEtran}
\bibliography{reference}

\end{document}